\documentclass[letterpaper]{article} 
\usepackage{aaai2026}  
\usepackage{times}  
\usepackage{helvet}  
\usepackage{courier}  
\usepackage[hyphens]{url}  
\usepackage{graphicx} 
\usepackage{natbib}  
\usepackage{caption} 
\usepackage{newfloat}
\usepackage{xspace}

\usepackage{listings}
\DeclareCaptionStyle{ruled}{labelfont=normalfont,labelsep=colon,strut=off} 
\usepackage{enumitem}
\usepackage{multirow}
\usepackage{soul}
\usepackage{pifont} 

\usepackage[T1]{fontenc}
\usepackage{multirow}

\usepackage{amsmath}
\usepackage{amssymb}
\usepackage{mathabx}
\usepackage{makecell}
\usepackage[dvipsnames]{xcolor}
\usepackage{color, colortbl}
\usepackage{booktabs}
\usepackage{caption}
\usepackage{listings}

\definecolor{citecolor}{HTML}{2980b9}
\definecolor{linkcolor}{HTML}{c0392b}
\definecolor{mycolor_blue}{HTML}{E7EFFA}
\definecolor{mycolor_green}{HTML}{E6F8E0}
\definecolor{mycolor_gray}{HTML}{ECECEC}
\definecolor{mycolor_red}{HTML}{FFE6E6}
\definecolor{mycolor_yellow}{HTML}{FFFFCC}
\definecolor{mycolor_purple}{HTML}{E6E6FF}
\definecolor{deepblue}{RGB}{0, 0, 200} 
\definecolor{brown}{RGB}{139, 69, 19} 
\definecolor{deepgreen}{RGB}{0, 100, 0}
\newcolumntype{R}{>{\color{red}}l}

\newcommand{\blue}[1]{\textbf{\textcolor{blue}{#1}}}

\newcommand{\eg}[1]{e.g.}

\newcommand{\ours}[1]{\textsc{DreamRunner}}
\newcommand{\attnmethod}[1]{SR3AI}
\newcommand{\dataname}[1]{DreamStorySet}

\def\onedot{\futurelet\@let@token\@onedot}
\def\@onedot{\ifx\@let@token.\else.\null\fi\xspace}

\def\ie{\emph{i.e}\onedot}

\title{\ours{}: Fine-Grained Compositional Story-to-Video Generation \\ with Retrieval-Augmented Motion Adaptation}

\author{Zun Wang$^1$
\qquad
Jialu Li$^1$
\qquad
Han Lin$^1$
\qquad
Jaehong Yoon$^2$
\qquad
Mohit Bansal$^1$
}
\affiliations{
$^1$UNC Chapel Hill \quad\quad $^2$Nanyang Technological University\\
\texttt{\{zunwang,jialuli,hanlincs,mbansal\}@cs.unc.edu}\quad\quad \texttt{jaehong.yoon@ntu.edu.sg}\\

\blue{\textbf{{\url{https://zunwang1.github.io/DreamRunner}}}}
}

\usepackage{bibentry}

\begin{document}


\twocolumn[{%
    \renewcommand\twocolumn[1][]{#1}%
    \maketitle
    \vspace{-25pt}
    \begin{center}
        \centering
            \includegraphics[width=\linewidth]{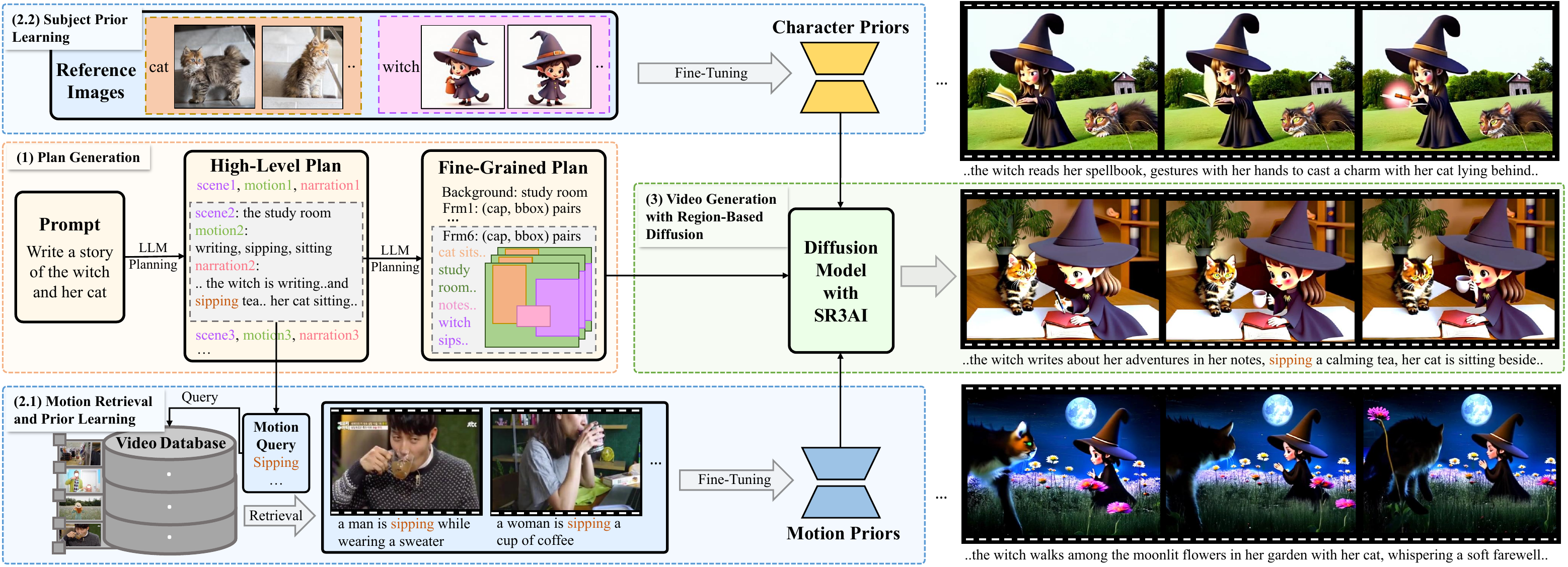}
        \captionof{figure}{
        \textbf{Overall pipeline for \ours{}.} \textbf{(1) plan generation stage:} we employ an LLM to craft a hierarchical video plan (\ie{}, ``High-Level Plan'' and ``Fine-Grained Plan'') from a user-provided generic story narration. \textbf{(2.1) motion retrieval and prior learning stage:} we retrieve videos relevant to the desired motions from a video database for learning the motion prior through test-time fine-tuning. \textbf{(2.2) subject prior learning stage:} we use reference images for learning the subject prior through test-time fine-tuning. \textbf{(3) video generation with region-based diffusion stage:} we equip diffusion model with a novel spatial-temporal region-based 3D attention and prior injection module (\ie{}, SR3AI) for video generation with fine-grained control. 
        }
        \label{fig:teaser}
        \vspace{2pt}
    \end{center}
}]

\begin{abstract}

Storytelling video generation (SVG) aims to produce coherent and visually rich multi-scene videos that follow a structured narrative. Existing methods primarily employ LLM for high-level planning to decompose a story into scene-level descriptions, which are then independently generated and stitched together. However, these approaches struggle with generating high-quality videos aligned with the complex single-scene description, as
visualizing such complex description involves coherent composition of multiple objects/events, complex motion synthesis
and character customization with sequential motions.
To address these challenges, we propose \textbf{\ours{}}, a novel story-to-video generation method: First, we structure the input script using a large language model (LLM) to facilitate both coarse-grained scene planning as well as fine-grained object-level layout planning. Next, \ours{} presents retrieval-augmented test-time adaptation to capture target motion priors for objects in each scene, supporting diverse motion customization based on retrieved videos, thus facilitating the generation of new videos with complex, scripted motions.
Lastly, we propose a novel spatial-temporal region-based 3D attention and prior injection module \attnmethod{} for fine-grained object-motion binding and frame-by-frame spatial-temporal semantic control. We compare \ours{} with various SVG baselines, demonstrating state-of-the-art performance in character consistency, text alignment, and smooth transitions. Additionally, \ours{} exhibits strong fine-grained condition-following ability in compositional text-to-video generation, significantly outperforming baselines on T2V-ComBench. Finally, we domonstrate \ours{}'s ability to generate multi-character interactions with qualitative examples. 
\end{abstract}
    
\section{Introduction}
\label{sec:intro}

Advancing storytelling video generation (SVG) is crucial for real-world video generation applications, enabling the creation of rich, immersive narratives with multiple realistic scenes, characters, and interactive events.
 Unlike existing short-form video generation approaches~\cite{singer2022make,hong2022cogvideo,
zhang2023controlvideo,chen2023control,wang2023modelscope,girdhar2023emu,khachatryan2023text2video,
weng2024art,qing2024hierarchical,zhang2024show,fei2024dysen,bar2024lumiere}, these models allow characters and objects to evolve across scenes, enhancing the coherence of generated content to align more closely with human storytelling. Such capabilities hold vast potential in media, gaming, and interactive storytelling.

Existing SVG methods~\cite{he2023animate,zhuang2024vlogger,oh2025mevg, zheng2024temporalstory, he2024dreamstory, zhao2024moviedreamer} primarily employ high-level planning with a large language model (LLM), breaking down a story into multiple key scene descriptions. Each scene is then generated independently as a separate video and later stitched together to form a complete long-form storytelling video. Generating high-quality single-scene video exhibits three key challenges: 1) \textit{Coherent composition}: As a highly complex textual form, a story, even at the single-scene level (e.g. ``\textit{Lucy on the left and a man on the right is walking towards each other, they meet in the middle and start ballroom dancing}''), typically involves multiple objects/characters with distinct motion trajectories, attributes, and sequentially occurring events, all of which must be coherently composed in the generated video. 2) \textit{Complex motion synthesis}: The complex scene descriptions often feature intricate character motions (e.g. ``\textit{ballroom dancing}'') that are difficult to generate from the base text-to-video (T2V) models. 3) \textit{Character customization with sequential events}: These descriptions usually involve characters with pre-defined reference images (e.g. \textit{Lucy}), with sequential motions (e.g. \textit{walking} to \textit{ballroom dancing}), making it challenging to maintain both temporal coherence and visual consistency with the character.
However, recent SVG methods often feed single-scene descriptions directly as textual conditions to the T2V model with limited constraints, resulting in suboptimal fidelity, missed events/objects, unclear motions, etc.

To address the above challenges, we propose \ours{}, a novel SVG framework that enhances fine-grained alignment between scene descriptions and generated videos. Beyond high-level planning, \ours{} uses LLM-based compositional reasoning to decompose complex scenes into frame-by-frame layout plans of multiple entities with sequential motions/events, followed by region-based attention for \textit{coherent composition}. For \textit{complex motion synthesis}, we adopt retrieval-augmented prior learning, injecting priors only into relevant regions to support \textit{character customization with sequential motions}. Specifically, 
\ours{} presents three essential processes in the framework: (1) \textit{Dual-Level Video Plan Generation}, (2)  \textit{Motion Retrieval and Subject/Motion Prior Learning},  and (3) \textit{Spatial-Temporal Region-Based 3D Attention and Prior Injection (SR3AI)}.
In \textbf{(1) plan generation stage}, given a user-provided story narration (e.g. ``\textit{write a story of the witch and her cat's one day}''), we employ an LLM for hierarchical planning: first generate a high-level plan with character-driven, motion-rich event descriptions across scenes, then decompose the scene descriptions into detailed, entity-specific frame-level layout plans within each scene.
The generated frame-level plan serves as the fine-grained guidance for T2V. 
In \textbf{(2) prior learning stage}, we learn both subject and motion priors to enhance character consistency and motion fidelity. Subject priors are learned from character reference images using customization techniques~\cite{ruiz2023dreambooth} to adapt the model to specific appearances. 
Then we treat \textit{complex motion synthesis} as a customization problem and learn motion priors to capture the visual patterns of target motions. To this end, we introduce an automatic retrieval pipeline that selects motion-relevant videos from a large-scale dataset~\cite{wang2023internvid} as references. We then apply test-time fine-tuning~\cite{zhao2023motiondirector} to learn customized motion priors. 
We use per-video prompts—rather than a shared one as in prior methods—to improve motion specificity, and learn both priors via LoRA-based tuning~\cite{hu2021lora} on specific layers of DiT~\cite{peebles2023scalable}.
In \textbf{(3) video generation stage}, we introduce \attnmethod{}, a novel spatial-temporal region-based 3D attention and prior injection module that enables fine-grained control without additional training.
Unlike prior methods that support only spatial~\cite{lin2023videodirectorgpt,lianllm,dav24,jain2024peekaboo} or temporal~\cite{bansal2024talc} control, \attnmethod{} leverages frame-level layout plans to enable spatial-temporal control over sequential events, object attributes, trajectories, and spatial relationships.
We first encode multiple conditions from the fine-grained plan. \attnmethod{} then computes visual latents for each condition based on its spatial-temporal layout and enforces attention masking so that each condition attends only to its designated region. This ensures precise control and coherent composition of multiple objects and motions.
Moreover, we extend this region-based design to inject learned character and motion priors into their corresponding regions in the diffusion model, enabling coherent character and motion customization.

We validate the effectiveness of \ours{} on two tasks: story-to-video generation and compositional text-to-video generation. For SVG, we collect a story dataset,  \dataname{}, and compare \ours{} with SoTA methods (VideoDirectGPT~\cite{lin2023videodirectorgpt} and VLogger~\cite{zhuang2024vlogger}). \ours{} achieves a 13.1\% relative improvement in character consistency score and an 8.56\% gain in text alignment score. It also improves sequential event generation within a single scene, with a 27.2\% boost for smoother multi-event transitions. Qualitative results further show strong generalization to multi-character settings.
In compositional T2V generation, \ours{} outperforms baseline methods on T2V-CompBench~\cite{sun2024t2v} across all dimensions, demonstrating its strength in compositional generation. Notably, despite being based on open-source models~\cite{yang2024cogvideox}, \ours{} achieves the highest scores in dynamic attribute binding and object interaction, along with comparable results in spatial relationships and motion binding to closed-source models, showing our method's potential to bridge the performance gap between open- and closed-source models.
In summary, our main contributions include:  
\begin{itemize}[noitemsep, topsep=4pt, leftmargin=*]
    \item A retrieval-augmented prior learning approach to enhance the synthesis of complex motions.
    \item A spatiotemporal region-based attention module for coherent composition of multiple objects and sequential events, along with a region-based LoRA injection design for character and sequential motion customization.
    \item SoTA performance in both compositional T2V and SVG.
\end{itemize}

\section{Related Work}
\noindent\textbf{Storytelling Video Generation.}
Storytelling video generation~\cite{oh2025mevg,zheng2024temporalstory,long2024videostudio} aims to produce multi-scene videos from input scripts. Existing approaches use either high-level LLM planning for step-by-step decomposition and generation~\cite{zhuang2024vlogger,he2023animate,lin2023videodirectorgpt}or keyframe generation with text-to-image models followed by video animation~\cite{he2024dreamstory,zhao2024moviedreamer,chai2023stablevideo}. Reference-based customization methods~\cite{ruiz2023dreambooth,TI23,custom_diffusion,StyleDrop,CAT24,StyleBreeder,ipadapter,ELITE,BLIPDiffusion23,chen2025multi,huang2025conceptmaster,liu2025phantom} to preserve character identity across scenes are also adopted.
Our work targets the video-centric challenge of generating multi-character, motion-rich videos with smooth, natural transitions.

\noindent\textbf{Compositional Generation.} Recent advances in diffusion models have enhanced compositional T2V generation by improving coherence, semantic alignment, and user control. Several methods leverage LLMs for scene planning~\cite{lianllm, lin2023videodirectorgpt, fei2024dysen, zheng2023layoutdiffusion,qu2023layoutllm}, while others employ regional masks for multi-object control~\cite{jain2024peekaboo, videoteris, dav24, yu2024zero, wei2024dreamvideo} or frame-level semantic conditioning~\cite{bansal2024talc, xing2024aid}. Additionally, LoRA-based compositional techniques integrate diverse concepts within the generation process~\cite{yang2024loracomposer, li2024selma, gu2024mix, zhong2024multi}. However, these approaches do not explicitly bind objects to their corresponding actions/events spatial-temporally. Our method ensures fine-grained control over both objects and motions, maintaining a cohesive object-action link throughout the video.

\noindent\textbf{Motion Customization.}
Motion customization remains a key challenge in video generation. One line of work focuses on pixel-level motion learning for video editing, aiming to replicate fine-grained motions from reference frames while preserving temporal consistency~\cite{wu2023tune, zhang2023motioncrafter, ren2024customize, jeong2023vmc, lin2024ctrl}. Another learns generic motion priors—e.g., human or camera movements—from curated datasets~\cite{zhao2023motiondirector, wu2023lamp, zhu2024echoreel, wei2023dreamvideo}, capturing high-level semantics for realistic motion synthesis. Most approaches rely on test-time fine-tuning with motion LoRAs or adapters.
Our method builds on this pipeline by retrieving motion-relevant videos from large-scale databases for more diverse and context-aware motion priors. We also replace the standard single-prompt conditioning with per-video detailed prompts to improve motion specificity and generation quality.
\begin{figure*}[t]
  \centering
    \includegraphics[width=0.97\textwidth]{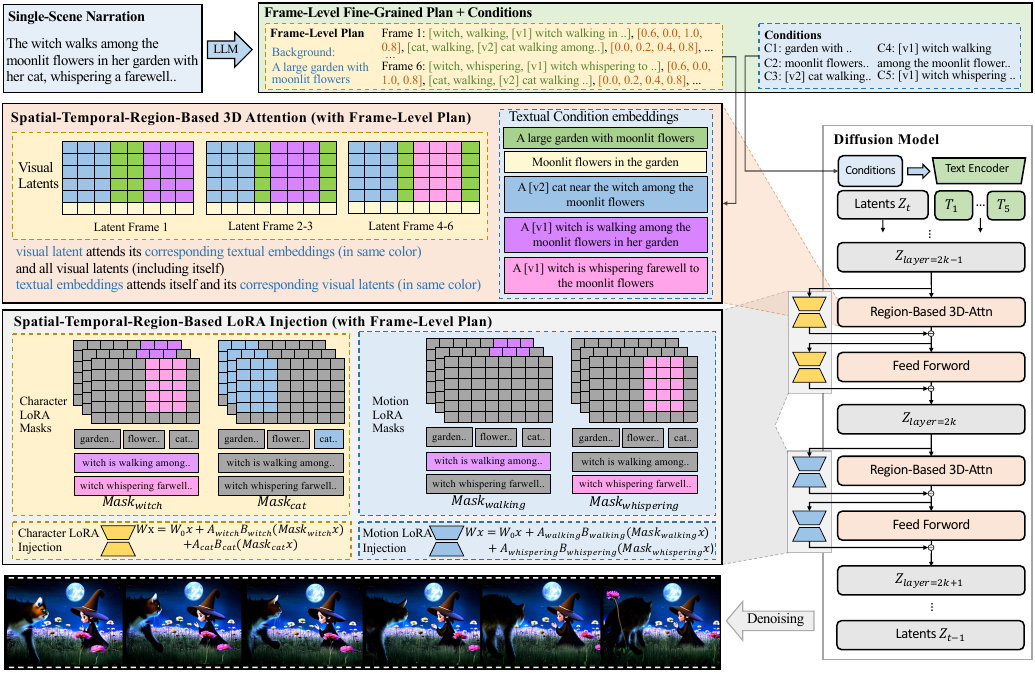}
  \caption{
  \textbf{Implementation details for region-based diffusion.} We extend the vanilla self-attention mechanism to {\sl spatial-temporal-region-based 3D attention} (see upper {\color{orange}orange} part), which is capable of aligning different regions with their respective text descriptions via region-specific masks. The region-based character and motion LoRAs (see lower {\color{Yellow}yellow} and {\color{cyan}blue} parts) are then injected interleavingly to the attention and FFN layers in each transformer block (see the right part).  Note that though we resize the visual latents into sequential 2D latent frames for better visualization, they are flattened and concatenated with all conditions when performing region-based attention. 
  Fig.~\ref{fig:mask} and Appendix~\ref{sr3i} provide example of the region-based attention mask and more details of region-based LoRA injection, respectively.}
  \label{fig:method}
\end{figure*}
\section{Methodology}
\noindent\textbf{Task Setup.}
Storytelling Video Generation focuses on creating multi-scene, character-driven videos based on a given topic. The characters are defined by reference images (\eg, images of a witch), and the topic is presented as an instructional prompt (\eg, "witch's one day"). The generated videos should align with the given topic and accurately reflect the characteristics and behavior of the characters.

\noindent\textbf{Method Overview.} Our approach employs a hierarchical system where an LLM generates event-based scripts across multiple scenes, followed by detailed plans specifying the layout and motion transitions of key objects per scene (Section~\ref{sec:subsec:plan}). A video diffusion model then synthesizes each scene step by step. We train motion priors from retrieval videos aligned with the LLM-generated plans, sourced from a large-scale video-language database, and character priors using the reference images (Section~\ref{sec:subsec:motion}). Finally, we inject these priors and detailed plans into the video generation process in a zero-shot manner using our spatial-temporal regional diffusion module \attnmethod{} (Section~\ref{sec:subsec:attn}).

\noindent\textbf{Base Generation Model.} We leverages CogVideoX-2B~\cite{yang2024cogvideox} as the base text-to-video model. CogVideoX-2B employs a DiT-based architecture that integrates full 3D attention, and generates 6-second videos at 8 fps conditioned on input text. In our method, we extend CogVideoX-2B by training character and motion priors in distinct layers (see Sec.~\ref{sec:subsec:motion}) and by modifying its 3D attention (see Sec.~\ref{sec:subsec:attn}) for better motion and character binding.

\subsection{Generating Dual-Level Plans with LLMs}\label{sec:subsec:plan}

\noindent\textbf{Story-Level Coarse-Grained Planning.} We prompt an LLM (GPT-4o~\citep{openai_gpt4o}) to generate 6$\sim$8 character-driven, motion-rich scene descriptions based on the story topic, task requirements, and a single in-context example. Each description follows a structured format: \texttt{\color{Purple}scene}, \texttt{\color{Green}motions}, and \texttt{\color{Red}narrations}, where motions are defined first, followed by corresponding event narrations. This sequence forms a high-level plan that guides story progression across scenes, ensuring narrative coherence.

\vspace{2pt}
\noindent\textbf{Scene-Level Fine-Grained Planning.}  
After generating a list of single-scene descriptions with narrations, we create detailed, entity-level plans for each latent frame. Each plan consists of an overall background description followed by entity-specific details for each latent frame.  
As shown in the yellow \textit{Frame-Level Plan} box at the top of Figure \ref{fig:method}, the background provides a global scene description (\eg{}, ``\textit{A large garden}"), formatted as \textcolor{deepblue}{\texttt{Background: background description}}. Entity-level details specify each entity’s description, motion (\eg{}, "\textit{A [v1] witch is walking among the moonlit flowers in her garden}"), and bounding box layout, formatted as: \texttt{Frame: \textcolor{deepgreen}{[entity name, entity motion, entity description]}, \textcolor{brown}{[x0,y0,x1,y1]}}. Here, \texttt{\textcolor{brown}{[x0,y0,x1,y1]}} denotes the top-left and bottom-right corners of the bounding box, with coordinates normalized to $[0,1]$. Entities without motion are labeled "\texttt{none}".  
When bounding boxes overlap, we prompt the LLM to generate a unified caption that integrates the descriptions of all entities within the overlapping region.
Each scene includes plans for six key frames, with each frame guiding one second of video generation (we interpolate key frames to match the \#frames of visual latents), resulting in a six-second output using CogVideoX. Detailed prompt templates for both levels' planning are in Appendix~\ref{llm prompts}.  

\subsection{Motion Retrieval and Prior Learning}\label{sec:subsec:motion}
\noindent\textbf{Retrieving Motion-Related Videos from Database.} 
We employ a retrieval-augmented approach to fine-tune motion priors at test time for complex motion synthesis. Based on motion descriptions generated from the LLM planning, we retrieve relevant videos from a large-scale video database~\cite{wang2023internvid}. Our retrieval pipeline first uses BM25~\cite{robertson2009probabilistic} for initial text-based retrieval, followed by attribute-based filtering and clip segmentation via object tracking~\cite{Jocher_YOLOv5_by_Ultralytics_2020}. We then compute semantic similarity scores using CLIP~\cite{radford2021learning} and ViCLIP~\cite{wang2023internvid} to refine the selection, ensuring high-quality motion-aligned videos (see Appendix \ref{retrival_process} for details). By following this process, we retrieve $4\sim20$ video clips per motion, which are then used as reference videos for learning motion priors.

\vspace{2pt}

\noindent\textbf{Motion Prior Training.}  
We follow recent motion customization methods~\cite{zhao2023motiondirector} with test-time fine-tuning for learning motion priors. Typically, reference videos are used to learn an appearance-debiased cross-video temporal LoRAs by injecting it into temporal attention layers with video-specific spatial LoRAs into spatial layers~\cite{wang2023modelscope, Zeroscope}.
Only temporal LoRAs are used during inference.
Since our approach is based on CogVideoX~\cite{yang2024cogvideox}, which employs 3D full attention instead of separate spatial-temporal attention, we manually designate even layers as ``spatial'' and odd layers as ``temporal'' to separate the learning of spatial and temporal LoRAs. We train LoRAs on top-ranked retrieved videos with all backbone parameters frozen, optimizing both diffusion loss for frame reconstruction and an appearance-debiased temporal loss to focus on motion-specific learning.
Note that unlike previous methods using a single-prompt condition for all the retrieved videos, we utilize video caption provided from the database as per-video prompt. This helps the model implicitly separate motion-unrelated backgrounds, appearances, etc, allowing it to focus on motion-specific patterns. More design details are in Appendix~\ref{motion prior training}.

\vspace{2pt}
\noindent\textbf{Subject Prior Learning.}
We learn the subject’s appearance by injecting LoRA modules into the spatial transformer layers. 
To train these LoRAs, we create videos by repeating reference images multiple times (48 time, similar to the output frame number of CogVideoX) and focus on reconstructing the first frame of the video during training, preventing overfitting to the static, repeated video. 
Notably, the subject priors are learned within spatial LoRAs, while the motion priors are learned within temporal LoRAs. Since their injections target different layers, there is no overlap, effectively avoiding conflicts between multiple LoRAs. 

\begin{table*}[!ht]
\setlength{\tabcolsep}{6pt} 
\renewcommand{\arraystretch}{1.2} 
\centering
\resizebox{1.0\linewidth}{!}{
\begin{tabular}{lcccccccccc}
\Xhline{1.0pt}
\multirow{2}{*}{\textbf{Method}} & \multicolumn{2}{c}{\textbf{Character}} & \multicolumn{2}{c}{\textbf{Fine-Grained Text}} &  \multicolumn{2}{c}{\textbf{Full Text}} & \multirow{1}{*}{\textbf{Transition}} & \multicolumn{3}{c}{\textbf{Visual Quality}} \\ 
\cmidrule(r){2-3}
\cmidrule(r){4-5}
\cmidrule(r){6-7}
\cmidrule(r){8-8}
\cmidrule(r){9-11}
 & CLIP & DINO & CLIP & ViCLIP & CLIP & ViCLIP & DINO & Aesthetics & Imaging & Smoothness \\ \hline
VideoDirectorGPT~\cite{lin2023videodirectorgpt} & 54.3 & 9.5 & 23.7   & 21.7 & 22.4 & 22.5 & 63.5 & 42.3 & 60.3 & 94.3 \\ 
VLogger~\cite{zhuang2024vlogger} & 62.5 & 41.3 & 23.5 & 23.1 & 22.5 & 22.2 & 73.6 & 43.4 & 61.2 & 96.2 \\ 
\multirow{2}{*}{\textbf{\ours{} (Ours)}} & 
\multirow{2}{*}{\begin{tabular}{c}
     \textbf{70.7}\vspace{-0.05in}\\
     \small{{\color{blue}(+13.1\%)}}
\end{tabular}} &
\multirow{2}{*}{\begin{tabular}{c}
     \textbf{55.1}\vspace{-0.05in}\\
     \small{{\color{blue}(+33.4\%)}}
\end{tabular}} &
\multirow{2}{*}{\begin{tabular}{c}
     \textbf{24.7}\vspace{-0.05in}\\
     \small{{\color{blue}(+5.11\%)}}
\end{tabular}} &
\multirow{2}{*}{\begin{tabular}{c}
     \textbf{23.7}\vspace{-0.05in}\\
     \small{{\color{blue}(+2.60\%)}}
\end{tabular}} &
\multirow{2}{*}{\begin{tabular}{c}
     \textbf{24.2}\vspace{-0.05in}\\
     \small{{\color{blue}(+7.56\%)}}
\end{tabular}} &
\multirow{2}{*}{\begin{tabular}{c}
     \textbf{24.1}\vspace{-0.05in}\\
     \small{{\color{blue}(+8.56\%)}}
\end{tabular}} &
\multirow{2}{*}{\begin{tabular}{c}
     \textbf{93.6}\vspace{-0.05in}\\
     \small{{\color{blue}(+27.2\%)}}
\end{tabular}} &
\multirow{2}{*}{\begin{tabular}{c}
     \textbf{55.4}\vspace{-0.05in}\\
     \small{{\color{blue}(+27.6\%)}}
\end{tabular}} &
\multirow{2}{*}{\begin{tabular}{c}
     \textbf{62.1}\vspace{-0.05in}\\
     \small{{\color{blue}(+1.47\%)}}
\end{tabular}} &
\multirow{2}{*}{\begin{tabular}{c}
     \textbf{98.1}\vspace{-0.05in}\\
     \small{{\color{blue}(+1.98\%)}}
\end{tabular}}
\\\\
\Xhline{1.0pt}
\end{tabular}
}
\vspace{-0.05in}
\caption{\textbf{Evaluation of story-to-video generation on \dataname{}.} We compare ours with VideoDirectorGPT and VLogger on character consistency (CLIP and DINO scores), text instructions following and full prompt adherence (CLIP and ViCLIP scores), and event transitions smoothness (DINO score). Our relative improvement over VLogger is highlighted in {\color{blue}blue}.}
\label{table:story_to_video_generation}
\end{table*}

\subsection{Spatial-Temporal-Region-Based Diffusion}\label{sec:subsec:attn}

\begin{figure}[t]
  \centering
  \includegraphics[width=1.0\linewidth]{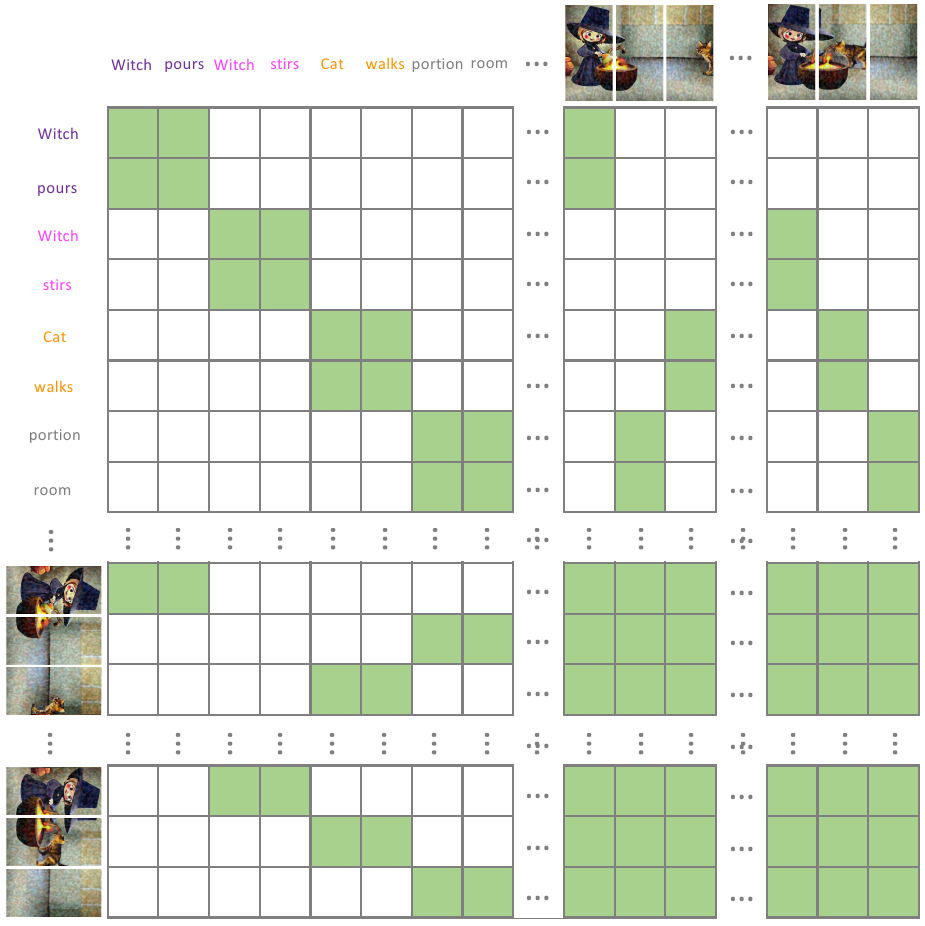}  
  \vspace{-0.2in}
  \caption{
  \textbf{Visualization of spatial-temporal region-based 3D attention mask}. Different text colors represent different conditions, while the white region indicates masked areas. For simplicity, we reduce each condition to two words, each frame to three segments, and display only three conditions and two frames in the figure. In practice, conditions can be longer and more numerous, frames can have more segments, and there are 12 latent frames in total.
  }
  \label{fig:mask}
\end{figure}

\paragraph{Region-Based 3D Attention.} 
We build our model on CogVideoX~\citep{yang2024cogvideox}, a text-to-video generation model designed on top of a Diffusion Transformer (DiT).
Unlike methods that use separate spatial and temporal attention for efficient video modeling, CogVideoX employs a 3D full attention module, integrating self-attention across concatenated embeddings of all visual latents and the text condition embeddings. We extend this module to enable region-specific conditioning via masking, aligning different regions with their respective text descriptions. 
Specifically, given a fine-grained plan with \( N \) region-specific text descriptions \( C_1, C_2, \dots, C_N \) and corresponding layouts \( L_1, L_2, \dots, L_N \) across frames, we encode each text condition \( C_i \) to produce embeddings \( T_1, T_2, \dots, T_N \) (Figure \ref{fig:method} top right). At each attention layer, we identify the visual latents corresponding to each layout \( L_i \) in the latent space. We then perform masked self-attention on the concatenation of \( T_1, T_2, \dots, T_N \) and \( L_1, L_2, \dots, L_N \). 
The self-attention mask is defined as follows: for each region's visual latents \( L_i \), attention is allowed to its corresponding text condition embeddings \( T_i \) and all visual latents \( L_1, L_2, \dots, L_N \). Conversely, for each condition embeddings \( T_i \), attention is restricted to itself and its corresponding latents \( L_i \). 
This design ensures each region is conditioned on its specific textual description while maintaining interactions among visual latents through unmasked attention among \( L_1, L_2, \dots, L_N \). 
No modifications are made to other modules in the base model, preserving the integrity of its original architecture. 
A visualization example of such masking strategy is contained in Fig.~\ref{fig:mask}.

\vspace{2pt}
\noindent\textbf{Region-Based LoRA Injection.}
We adopt a similar region-based strategy for injecting LoRA priors into diffusion models. For each LoRA, we first identify the corresponding regions of latent tokens based on the associated text description and layout information. LoRA injection is then applied exclusively to these regions, ensuring precise alignment between the priors and their designated areas. This approach enables handling multiple LoRAs simultaneously while avoiding conflicts between them, preserving the integrity of each injected prior. Appendix~\ref{sr3i} provides details of this strategy with equation derivations, explanations, etc.

\begin{table*}[t]
\resizebox{0.9\textwidth}{!}{%
\begin{tabular}
{lllllll}
\toprule 
\multicolumn{1}{l}{{\textbf{Model}}} & \multicolumn{1}{l}{\textbf{Consist-attr}} & \multicolumn{1}{l}{\textbf{Dynamic-attr}} & \multicolumn{1}{l}{\textbf{Spatial}} & \multicolumn{1}{l}{\textbf{Motion}} & \multicolumn{1}{l}{\textbf{Action}} & \multicolumn{1}{l}{\textbf{Interaction}} \\
\midrule
\rowcolor{gray!10} 
Gen-3~\cite{gen3} & 0.7045 &  0.2078 & 0.5533 & \cellcolor{mycolor_yellow}{\emph{0.3111}} & 0.6280 & 0.7900 \\
\rowcolor{gray!10} 
Dreamina~\cite{Dreamina} & \cellcolor{mycolor_yellow}{\emph{0.8220}} &  0.2114 & 0.6083 & 0.2391 & 0.6660 & 0.8175  \\
\rowcolor{gray!10} 
PixVerse~\cite{PixVerse} & 0.7370 &  0.1738 & 0.5874 & 0.2178 & \cellcolor{mycolor_yellow}\emph{0.6960} & 0.8275 \\
\rowcolor{gray!10} 
Kling~\cite{kling} & 0.8045 & \cellcolor{mycolor_yellow}{ \emph{0.2256}} & \cellcolor{mycolor_yellow}{\emph{0.6150}} & 0.2448 & 0.6460 & \cellcolor{mycolor_yellow}{\emph{0.8475}} \\
\midrule
VideoCrafter2~\cite{chen2024videocrafter2} &0.6750 &0.1850 & 0.4891& 0.2233&0.5800 &0.7600 \\
Open-Sora 1.2~\cite{opensora} &0.6600 &0.1714 & 0.5406& 0.2388&0.5717 &0.7400 \\
Open-Sora-Plan v1.1.0~\cite{pku_yuan_lab_and_tuzhan_ai_etc_2024_10948109} &\underline{0.7413} &0.1770 & 0.5587 &0.2187& \textbf{0.6780}&0.7275 \\
VideoTetris~\cite{videoteris} & 0.7125&0.2066 & 0.5148& 0.2204&0.5280 & 0.7600 \\
LVD~\cite{lian2023llm} &0.5595 &0.1499 & 0.5469& \underline{0.2699}&0.4960 &0.6100 \\

\hline
CogVideoX-2B~\cite{yang2024cogvideox} & 0.6775 &  0.2118 & 0.4848 & 0.2379 & 0.5700 & 0.7250 \\
\textbf{CogVideoX-2B+SR3A (Ours)} & 0.7350 \small{{\color{blue}(+8.5\%)}} &  \underline{0.2672} \small{{\color{blue}(+26.2\%)}} & \underline{0.6123} \small{{\color{blue}(+26.3\%)}} & 0.2608 \small{{\color{blue}(+9.6\%)}} & 0.5840 \small{{\color{blue}(+2.5\%)}} & \underline{0.7625} \small{{\color{blue}(+5.2\%)}} \\
\hline
CogVideoX-5B~\cite{yang2024cogvideox} & 0.7232 &  0.2250 & 0.5845 & 0.2551 & 0.6040 & 0.7995 \\
\textbf{CogVideoX-5B+SR3A (Ours)} & \textbf{0.7650} \small{{\color{blue}(+5.8\%)}} &  
\textbf{0.2832} \small{{\color{blue}(+25.9\%)}} & 
\textbf{0.6875} \small{{\color{blue}(+17.5\%)}} &  
\textbf{0.3041} \small{{\color{blue}(+19.2\%)}} &  
\underline{0.6340} \small{{\color{blue}(+5.0\%)}} &  
\textbf{0.8725} \small{{\color{blue}(+9.1\%)}} \\
\bottomrule
\end{tabular}
}\centering
\vspace{-0.1in}
    \caption{\textbf{T2V-CompBench evaluation results}. Best/2nd best
scores for open-sourced models are \textbf{bolded}/\underline{underlined}. 
\colorbox{mycolor_gray}{gray} indicates close-sourced models, and \colorbox{mycolor_yellow}{\emph{yellow}} indicates the best score for close-sourced models. 
}
\vspace{-0.1in}
\label{tab:benchmark}
\end{table*}

\begin{table}[t]
\setlength{\tabcolsep}{3pt} 
\renewcommand{\arraystretch}{1.2} 
\centering
\resizebox{1.0\linewidth}{!}{
\begin{tabular}{cccccccccc}
\Xhline{1.0pt}
\multirow{2}{*}{\textbf{RAG}} & \multirow{2}{*}{\textbf{SR3AI}} & \multicolumn{2}{c}{\textbf{Fine-Grained Text}} &  \multicolumn{2}{c}{\textbf{Full Text}} & \multirow{1}{*}{\textbf{Trans.}} &  \multicolumn{3}{c}{\textbf{Quality}} \\ 
\cmidrule(r){3-4}
\cmidrule(r){5-6}
\cmidrule(r){6-7}
\cmidrule(r){8-10}
& & CLIP & ViCLIP & CLIP & ViCLIP & DINO & Asth. & Img. & Smth.  \\ \hline
$\times$& $\times$ &  23.8   & 22.5 & 22.2 & 22.1 & 87.1 & 54.3 & 61.3 & 94.3 \\ 
$\times$ & \checkmark &  23.9   & 23.1 & 23.5 & 22.4 & 92.5 & 55.4 & 61.9 & 98.0 \\ 
\checkmark &  $\times$ & 24.7 & 23.5 & 23.9 & 24.0 & 84.6 & \textbf{55.6} & 61.9 & 98.1 \\ 
\checkmark & \checkmark & \textbf{24.7} & \textbf{23.7} & \textbf{24.2} & \textbf{24.1}  & \textbf{93.6} & 55.4 & \textbf{62.1} & \textbf{98.1} \\ 
\Xhline{1.0pt}
\end{tabular}
}
\caption{\textbf{Ablation studies for the effectiveness of RAG and \attnmethod{} in \ours{}.}  Our full model achieves the best text-following ability and event transition smoothness.
}
\label{tab: ablations}
\end{table}

\section{Experiments}
In this section, we first introduce the evaluation datasets and evaluation metrics details in Section \ref{subsec:evaluation_dataset_metrics}, then compare our \ours{} with prior methods on story-to-video generation in  Section \ref{subsec:story_to_video_generation}. Next, we present detailed ablation studies on the necessity of RAG and effectiveness of \attnmethod{} in Section \ref{subsec:ablation_studies}, and demonstrate the generalizability of our \ours{} to improve compositional text-to-video generation on T2V-CompBench~\cite{sun2024t2v} in Section \ref{subsec:t2v_compbench_evaluation}. Moreover, we show the effectiveness of RAG for learning the motion prior on a more comprehensive motion dataset we collect in Section \ref{subsec:rag_motion_prior}. Lastly, we present qualitative comparison between our \ours{} and previous approaches in Section \ref{subsec:qual}.

\subsection{Experimental Setups}
\label{subsec:evaluation_dataset_metrics}

\textbf{Evaluation Datasets.} We evaluate \ours{} on two tasks: (1) story-to-video generation, and (2) compositional text-to-video generation. The first task focuses on the model's ability to follow the text closely while maintaining character and scene consistency throughout the story. The second task assesses various aspects of compositionality in video generation. For (1) story-to-video generation, we collect and introduce a new benchmark dataset, \dataname{}. Specifically, we collect 10 characters, including 6 from existing customization datasets  (CustomConcept101~\citep{kumari2023multi} and Dreambooth~\citep{ruiz2023dreambooth}), and 4 with generation models (FLUX~\citep{flux}). (featuring two motions per scene) and three multi-character stories (featuring two or three motions per scene). Each story comprises 5 to 8 scenes, incorporating a total of 64 diverse motions throughout. We focus on single-character stories for quantitative evaluation of SVG models and reserve multi-character stories for qualitative evaluation. For (2) compositional text-to-video generation, we use the T2V-CompBench~\citep{sun2024t2v} to benchmark the performance of \ours{}, where we select six dimensions except numeracy.

\noindent\textbf{Evaluation Metrics.}
We evaluate our storytelling videos across multiple dimensions: Character Consistency (Frame-to-Reference CLIP/DINO), Text Alignment at both the full narration and fine-grained scene level (Image/Video-to-Text CLIP/ViCLIP), and Transition Smoothness (Frame-to-Frame DINO). For visual quality, we adopt three representative metrics from VBench~\cite{huang2023vbench}: aesthetic quality, imaging quality, and video smoothness~\cite{li2023amt}, with full quality results and metric details provided in the Appendix.
We follow similar evaluation metrics to T2V-ComBench~\cite{sun2024t2v} for compositional T2V.

\noindent\textbf{Implimentations.} We use CogVideoX-2B as our base model for SVG. Test-time-finetuning each prior requires 5min on a single A6000 GPU. For compositional T2V we evaluate with both CogVideoX-2B and CogVideoX-5B.

\subsection{Story-To-Video Generation Evaluation}
\label{subsec:story_to_video_generation}

We compare \ours{} with prior SoTAs (VideoDirectorGPT~\cite{lin2023videodirectorgpt} and VLogger~\cite{zhuang2024vlogger}) on our \dataname{} dataset for story-to-video generation. For fairness, each scene narration is split into two single-motion descriptions, with corresponding videos later merged into a single-scene video. As shown in Table~\ref{table:story_to_video_generation}, \ours{} improves CLIP/DINO scores by 13.1\%/33.4\% over VLogger, demonstrating the effectiveness of our learned subject prior and region-based LoRA injection for character consistency. To evaluate text-following capability, we assess both full-prompt adherence and fine-grained event alignment. \ours{} improves CLIP/ViCLIP scores consistently on both settings, showing superior alignment with both full-scene and fine-grained event descriptions. For transition quality, we compute the DINO-based transition score to measure scene and event consistency. \ours{} improves transitions by 27.2\% over VLogger, highlighting the effectiveness of \attnmethod{} in generating sequential events in a single scene. Lastly, we evaluate visual quality across aesthetic quality, imaging quality, and motion smoothness. \ours{} enhances aesthetics while slightly improving the other two, demonstrating its capability to generate high-quality videos adhere to complex scene descriptions with smooth event transitions.
We provide three additional quality scores from VBench and qualitative examples in the Appendix.

\begin{figure*}[t]
  \centering
    \includegraphics[width=1.0\textwidth]{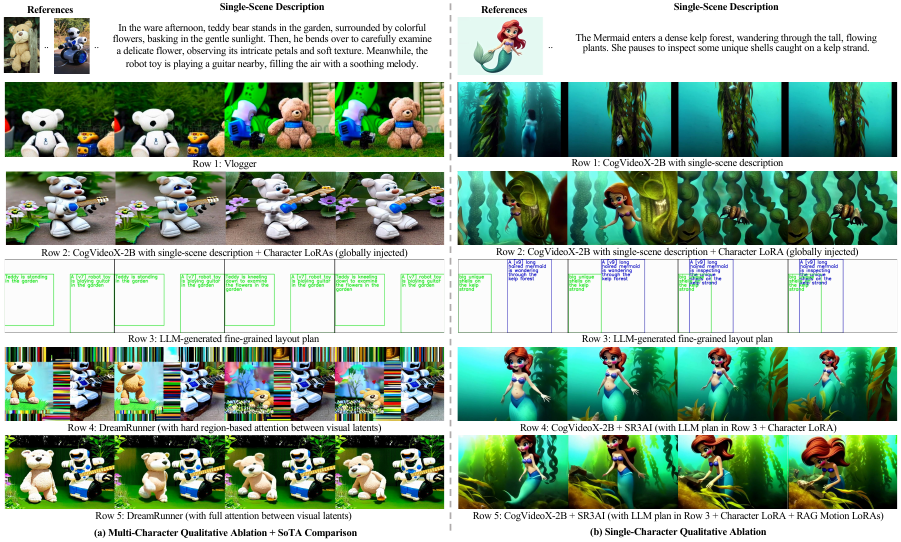}
    \vspace{-7mm}
  \caption{
  \textbf{Qualitative comparison and ablations of \ours{} on SVG.} In (a) multi-character example, \ours{} produces significantly better character consistency compared to other strong baselines, while others fail to maintain object consistency (e.g., VLogger), or fail to generate multiple objects ((a) Row 2,4). In (b) single-character setting, integrating \attnmethod{} and locally-injected priors consistently improve overall quality, complex motion synthesis and coherent composition. Note that in the overlapped regions in (b) row 3, the caption is a merge of the two. For cleaner visualization, we don’t show it here.
  }
  \label{fig:qual}
\end{figure*}

\subsection{Ablation Studies}
\label{subsec:ablation_studies}
In this section, we demonstrate the effectiveness of RAG for automatic video retrieval in motion prior learning and \attnmethod{} for fine-grained control over objects and their motions to achieve coherent composition. As shown in Table~\ref{tab: ablations}, using \attnmethod{} for enhanced multi-object multi-event binding (2nd row) significantly improves event transition smoothness within a single scene, as well as visual quality and text alignment—likely because decomposing attention into region-based components enables the model to “divide and conquer” more effectively. Incorporating retrieval-augmented motion prior learning (3rd row) further improves video-text similarity for both fine-grained and full-prompt alignment, showing its importance for motion enhancement. Finally, combining both(last row) yields the best results across event transition, text alignment, and visual quality. We also provide additional ablations on RAG pipeline, Layer-seperation strategy for motion prior learning, and computational cost comparison in Appendix~\ref{appendix ablation}.

\subsection{Qualitative Ablations and Comparisons}
\label{subsec:qual}

We provide qualitative comparisons and ablations in Fig.~\ref{fig:qual}.

In Fig.~\ref{fig:qual}(a), we compare \ours{} with VLogger\cite{zhuang2024vlogger} for multi-character generation and analyze the effects of region-based attention and LoRA injection. Our method (row 5) generates coherent multi-character composition and motions, outperforming VLogger (row 1). Using CogVideoX-2B with  character LoRAs injected globally (row 2) results in interference, producing a robot-like teddy bear and blurry compositions, highlighting the necessity of region-based LoRA injection.
To study attention design, we ablate our region-based attention by comparing hard-regional attention (row 4) and full attention (row 5). While hard-regional attention strictly follows the layout plan (row 3), it limits spatial-temporal continuity due to lack of inter-region interaction. In contrast, our full attention mechanism enables smooth transitions while maintaining spatial-region constraints, supporting high-quality multi-character customization.

In Fig.~\ref{fig:qual}(b), we present single-character ablations. Using the base CogVideoX-2B with only scene-level text (row 1) leads to vague character/background and missing actions. Injecting global character LoRA (row 2) improves character appearance but still fails on action and transition quality, and degrades background fidelity (e.g., cartoonish kelp forest). Applying \attnmethod{} with layout plans (row 4) improves trajectory control and preserves background fidelity through localized injection, but motion remains limited. Injecting RAG-learned motion priors (row 5) enables clear, fine-grained motion execution (e.g., the mermaid stooping and interacting with kelp), demonstrating the benefit of our motion prior learning and injection strategy.
Overall, our full model combines coherent composition with strong motion quality, showing the effectiveness of \attnmethod{} and retrieval-augmented prior learning for complex, multi-entity video generation.

\begin{table}[t]
\setlength{\tabcolsep}{6pt} 
\renewcommand{\arraystretch}{1.2} 
\centering
\resizebox{0.48\textwidth}{!}{
\begin{tabular}{c|cc}
\Xhline{1.0pt}
\textbf{Method} & \textbf{CLIP} & \textbf{ViCLIP } \\ \hline
CogVideoX-2B & 23.39 & 20.84 \\ \hline
CogVideoX-2B + RAG (w/ single prompt for all videos) & 24.01 & 22.02 \\ 
CogVideoX-2B + RAG (w/ per-video prompt) & \textbf{24.67} & \textbf{23.04} \\ \Xhline{1.0pt}
\end{tabular}
}
\vspace{-0.1in}
\caption{{\textbf{Effect of RAG and per-video prompt for motion prior learning}.}
\vspace{-0.15in}
}
\label{tab:motionprior}
\end{table}

\subsection{Compositional T2V Generalization}
\label{subsec:t2v_compbench_evaluation}
In this section, we demonstrate how our spatial-temporal region-based attention module (SR3A) enhances compositional T2V, as evaluated on T2V-CompBench~\cite{sun2024t2v}. We use SR3A (no LoRA injection) as no customization is required. Given a prompt, we use GPT-4o to generate layout plans, and SR3A ensures coherent composition of objects and events. As shown in Table~\ref{tab:benchmark}, SR3A significantly improves both CogVideoX-2B and CogVideoX-5B~\cite{yang2024cogvideox} across all categories. Specifically, it boosts dynamic attribute binding by over 25\%, spatial binding by over 15\%, and motion binding by at least 10\%, highlighting SR3A’s ability to maintain coherent multi-object compositions, trajectories, and sequential events. It also improves scores on other fine-grained aspects, demonstrating strong control capabilities. Notably, \ours{} built on CogVideoX-5B achieves SoTA results in five dimensions among open-source models, and surpasses all closed-source models in dynamic attribute binding, spatial binding, and object interactions, highlighting its ability to close the open-closed source model gap and adapt to stronger base models. Qualitative examples per dimension are in Appendix~\ref{t2vcomqual}.

\subsection{Effect of RAG and Per-Caption Prompt} 
\label{subsec:rag_motion_prior}
We investigate the effectiveness of retrieval-augmented test-time fine-tuning and our per-caption prompt design for learning an enhanced motion prior. Specifically, for each motion in the 64-motion set, we use GPT-4o to generate six prompts and evaluate the average CLIP/ViCLIP scores. As shown in Table~\ref{tab:motionprior}, applying our approach to CogVideoX-2B improves both scores, with the significant ViCLIP gain indicating better story-video alignment and enhanced motion accuracy. Rows 2–3 further show that per-video prompts outperform single prompts, suggesting that video-specific conditioning helps the model ignore unrelated visual cues and better capture motion-specific patterns. These results confirm that RAG effectively retrieves motion-relevant videos and facilitates the learning of more accurate motion priors.

\section{Conclusion}
In this work, we present \ours{}, a novel framework for story-to-video generation. Specifically, \ours{} utilizes a LLM to structure a hierarchical video plan, then introduces retrieval-augmented test-time adaptation to capture target motion priors, and finally generates videos using a novel region-based 3D attention and prior injection module for coherent composition.
Experiments on both story-to-video and compositional T2V generation benchmarks show that \ours{} outperforms strong baselines and SoTAs in tackling fine-grained complex motions, maintaining multi-scene consistency of multiple objects, and ensuring seamless scene transitions.

\section*{Acknowledgments}
This work was supported by DARPA ECOLE Program No. HR00112390060, NSF-AI Engage Institute DRL2112635, DARPA Machine Commonsense (MCS) Grant N66001-19-2-4031, ARO Award W911NF2110220, ONR Grant N00014-23-1-2356, Accelerate Foundation Models Research program, and a Bloomberg Data Science PhD Fellowship. The views contained in this article are those of the authors and not of the funding agency.

\small
\bibliography{aaai2026}


\appendix

\section{Additional Method Details}

\subsection{Retrieval Processes}
\label{retrival_process}
\noindent Our retrieval process consists of the following steps (using the example “\textit{sitting}” as the query motion): 
\begin{itemize}[noitemsep, topsep=4pt, leftmargin=*]
\item[1)]  Initial Retrieval with BM25: We use the text-only BM25 score~\cite{robertson2009probabilistic} based on video captions in the dataset to retrieve 400 candidate videos for the query. To ensure the retrieved videos are human motion-centric, we add "\textit{person is}" at the beginning of the query ("\textit{person is sitting}").

\item[2)] Attribute-Based Filtering: We refine the candidate pool by filtering videos based on key attributes such as duration (at least 2 second), frame count (at least 40 frames), and aspect ratio (width/height at least 0.9).
This ensures that the selected videos align with the requirements of the video generator, excluding videos that are too short or have extreme aspect ratios.

\item[3)] Clip Segmentation via Object Tracking: We track individuals within videos using YOLOv5~\citep{Jocher_YOLOv5_by_Ultralytics_2020} and segment clips into human-centered segments based on the tracking results, keeping meaningful human-focused content.

\item[4)] Scoring with CLIP and ViCLIP~\citep{radford2021learning, wang2022internvideo}: To ensure the fidelity between the segmented video clips and the query, we compute semantic similarity scores to the query text (\eg, “the person is sitting”) using CLIP and ViCLIP for each segmented clip. The CLIP score is computed by sampling eight frames and averaging frame-query scores, while the ViCLIP score is directly computed on the full video and query. We select the top 20 videos that satisfy the average scores of CLIP and ViCLIP > 0.2. We retain the top four videos based on their ranking if fewer than four videos meet this threshold.
\end{itemize}

\subsection{Motion Prior Training}
\label{motion prior training}
We train the LoRAs on our filtered top-ranked videos with all other backbone parameters frozen, using two diffusion losses: a standard diffusion loss $L_{org}$, which is a reconstruction loss of all the video frames, and an appearance-debiased temporal loss $L_{ad}$, which decouples the motion space from appearance space in the latent space, focusing on only reconstructing the motions in the videos. Formally, 
\begin{equation}
L_{org} = E_{z_0, y, \epsilon \sim N(0,1), t \sim U(0, T)}[||\epsilon - \epsilon_{\theta}(z_t, t, y)||_2]
\end{equation}
where $z_0$ is the latent encoding of the training videos, $y$ is the text prompt condition, $\epsilon$ is the Gaussian noise added to the latent space, $\epsilon_\theta$ is the predicted noise, and $t$ is the denoising time step. The appearance-debiased temporal loss optimizes the normalized latent space:
\begin{equation}
\phi (\epsilon) = \sqrt{\beta^2+1}\epsilon - \beta\epsilon_{anchor}
\end{equation}
where $\epsilon_{anchor}$ is the anchor among the frames from the same training data, and $\beta$ is the strength factor that controls the strength of the debiasing. $L_{ad}$ is defined as:
\begin{equation}
L_{ad} = E_{z_0, y, \epsilon \sim N(0,1), t \sim U(0, T)}[||\phi(\epsilon) - \phi(\epsilon_{\theta}(z_t, t, y))||_2]
\end{equation}
In the end, we update the model using a combined motion loss function defined as $L_{motion} = L_{org} + L_{ad}$. Notably, we do not apply scaling to each loss term throughout experiments in the paper, highlighting the robustness and simplicity of hyperparameter selection in \ours{}.

\subsection{Region-Based LoRA Injection}
\label{sr3i}
Low-Rank Adaptation (LoRA)~\citep{hu2021lora} was introduced to efficiently adapt large pre-trained language models to various downstream tasks. More recently, LoRA has been extended to text-to-image and text-to-video generation, enabling lightweight subject and motion customization~\citep{ruiz2023dreambooth, zhao2023motiondirector}. Instead of updating the full weight matrix $W$, LoRA applies a low-rank decomposition, modifying it as:  
\begin{equation}
W = W_0 + \Delta W = W_0 + BA,
\end{equation}  
where $W_0\in \mathbb{R}^{d\times k}$ denotes the original model weights, and $B\in \mathbb{R}^{d\times r}$, $A\in \mathbb{R}^{r\times k}$ are the learnable low-rank matrices with $r \ll d,k$. 

During inference, the LoRA layer's output is the summation of the original layer's output $W_0x$ and low-rank matrices' output $BAx$:
\begin{equation}
Wx = W_0x + BAx
\end{equation}  
where  $x\in \mathbb{R}^{ k\times c}$ ($c$ is the number of latent tokens, while $k$ is the dimension of each token) is the latents.
By restricting updates to these low-rank factors, LoRA significantly reduces computational overhead compared to full fine-tuning approaches. Additionally, its plug-and-play nature makes it highly efficient for deployment and sharing across different pre-trained models.

To inject LoRA locally into specific regions, we compute masks for the latents corresponding to each LoRA and mask the input of other regions. For example, considering the character LoRAs in main paper's Fig. 1, we compute the output when forwarding the latents $x$ to this layer as:
\begin{equation}
\begin{aligned}
Wx &= W_0x + A_{\text{witch}} B_{\text{witch}}(Mask_{\text{witch}}\cdot x) 
\\ &+ A_{\text{cat}} B_{\text{cat}} (Mask_{\text{cat}}\cdot x) \\
&= (W_0 + A_{\text{witch}} B_{\text{witch}})( Mask_{\text{witch}} \cdot x) 
\\ &+ (W_0+ A_{\text{cat}} B_{\text{cat}})(Mask_{\text{cat}} \cdot x)
\\ &+ W_0((\mathbf{1}_{c} - Mask_{\text{witch}} - Mask_{\text{cat}})\cdot x)
\end{aligned}
\end{equation}
where $\mathbf{1}_{c}$ is a $c$ dimension all-ones vector and $\cdot$ denotes the dot product, so $Mask \cdot x$ denotes the dot product between the $c$ dimension boolen mask and the $k\times c$ latent. In this case, the cat's LoRA only acts on the cat-related region ($Mask_{\text{cat}} \cdot x$), ensuring that the cat-region output is similar to the output obtained when applying a classical cat LoRA globally:
\begin{equation}
\begin{aligned}
 & Wx_{Mask_{\text{witch}}=1} \\
=& (W_0 + A_{\text{witch}} B_{\text{witch}})( Mask_{\text{witch}} \cdot x_{Mask_{\text{witch}}=1}) 
\\ +& (W_0+ A_{\text{cat}} B_{\text{cat}})(Mask_{\text{cat}} \cdot x_{Mask_{\text{witch}}=1})
\\ +& W_0((\mathbf{1}_{c} - Mask_{\text{witch}} - Mask_{\text{cat}})\cdot x_{Mask_{\text{witch}}=1}) \\
=& (W_0 + A_{\text{witch}} B_{\text{witch}})( \textbf{1} \cdot x_{Mask_{\text{witch}}=1}) 
\\ +& (W_0+ A_{\text{cat}} B_{\text{cat}})( \textbf{0} \cdot x_{Mask_{\text{witch}}=1})
\\ +& W_0( \textbf{0} \cdot x_{Mask_{\text{witch}}=1}) \\
=& (W_0 + A_{\text{witch}} B_{\text{witch}})x_{Mask_{\text{witch}}=1}
\end{aligned}
\end{equation}
Similarly, the witch's LoRA only modifies the witch-related region ($Mask_{\text{witch}} \cdot x$), preserving consistency with traditional LoRA-based adaptation for the witch.  
For other regions unrelated to both the witch and the cat, the output remains unchanged, as the mask ensures that no LoRA modifications are applied, making it equivalent to the output without any LoRA injection.
For the Motion LoRAs in main paper's Fig. 1,, we follow the similar style to inject them locally:
\begin{equation}
\begin{aligned}
    Wx = W_0x +  A_{\text{walking}} B_{\text{walking}} (Mask_{\text{walking}} \cdot x) \\
    + A_{\text{whispering}} B_{\text{whispering}} (Mask_{\text{whispering}} \cdot x)
\end{aligned}
\end{equation}
Such a region-based LoRA injection design allows multiple LoRAs to act on different regions, ensuring seamless multi-LoRA integration while preventing interference between unrelated entities or motions, thereby enabling precise multi-character, multi-motion customization.

\section{Additional Comparisons}
\label{appendix ablation}

\subsection{Ablations of RAG pipeline}
\label{rag ablation}
We provide additional experiments in Section 3.2 to demonstrate the effectiveness of our data processing approach within the retrieval pipeline. Table \ref{tab:motionprior_ablation} primarily evaluates the impact of the maximal number of retrieved videos and the use of CLIP and ViCLIP for filtering. For efficiency, we evaluate a subset of eight motions selected from the full pool of 64 motions.
The results indicate that, compared to the CogVideoX zero-shot baseline (Row 1), retrieving videos without any filtering (Row 2) improves performance, even though BM-25~\cite{robertson2009probabilistic} retrieval introduces some noise. This highlights the importance of retrieval itself. Furthermore, adding CLIP and ViCLIP as filters (Row 4) further enhances performance, showcasing the benefit of using semantically aligned videos for improved motion learning. Additionally, retrieving a sufficient number of videos is critical, as evidenced by Row 3, where limiting retrieval to a maximum of three videos results in poorer performance compared to retrieving 20 videos (Row 4).
\begin{table}[h!]
\setlength{\tabcolsep}{6pt} 
\renewcommand{\arraystretch}{1.2} 
\centering
\resizebox{0.9\linewidth}{!}{
\begin{tabular}{cc|cc}
\Xhline{1.0pt}
\textbf{Max. \#Retrieval} & \textbf{CLIP+ViCLIP filter} & \textbf{CLIP} & \textbf{ViCLIP } \\ \hline
0 & $\times$& 23.42 & 20.56 \\ \hline
20 & $\times$ & 24.01 & 22.51 \\ \hline
3 & \checkmark & 24.45 & 22.80 \\ \hline
20 & \checkmark & \textbf{25.47} & \textbf{23.66} \\ 
\Xhline{1.0pt}
\end{tabular}
}
\vspace{-0.1in}
\caption{{Pipeline component ablation on retrieval-augmented test-time adaptation for learning a better motion prior.
}
\vspace{-0.15in}
}
\label{tab:motionprior_ablation}
\end{table}

\subsection{Ablations of Odd-Even Layer Separation for Motion Prior Learning}
\label{lora design ablation}

During motion prior learning, since our backbone does not explicitly separate spatial and temporal modeling, we manually designate even layers as spatial and odd layers as temporal to decouple the learning of spatial and temporal LoRAs.  
We hypothesize that this design helps learn appearance-debiased, motion-focused priors, while the interleaved structure allows adaptation across low-level structures and high-level semantics, as both spatial and temporal LoRAs are injected throughout the model. We validate this hypothesis in Table~\ref{tab:ablation2}. Specifically, removing spatial LoRAs (row 1) degrades performance compared to row 3, indicating that separating spatial and temporal LoRAs enhances motion-focused learning. Additionally, the Half-Half approach (injecting spatial LoRAs into the first half of layers and temporal LoRAs into the second half, row 2) underperforms compared to interleaved injection, highlighting the importance of multi-level learning across the entire model.

\begin{table}[h]
\setlength{\tabcolsep}{6pt} 
\renewcommand{\arraystretch}{1.2} 
    \centering
    \resizebox{0.7\linewidth}{!}{
    \begin{tabular}{l|cc}
    \Xhline{1.0pt}
    \textbf{Method} & \textbf{CLIP} & \textbf{ViCLIP} \\ \hline
    No Appearance-Debiased & 24.4 & 22.5 \\ \hline
    Half-Half Injection & 24.9 & 23.1 \\ \hline
    Interleaved Injection (Ours)& \textbf{25.5} & \textbf{23.7} \\ \hline
    \Xhline{1.0pt}
    \end{tabular}
    }
    \vspace{-0.1in}
    \caption{LoRA injection ablations.}
    \label{tab:ablation2}
\vspace{-1.45em}
\end{table}

\subsection{Computational Cost Comparison}

Since our method is training-free and only involves test-time fine-tuning (TTF), we report the optimization time solely for each method for reference. Specifically, our methods' TTF takes approximately 0.2 GPU hours per motion, totaling 3–4 GPU hours for a full story with 15–20 motions.

In comparison, training-intensive baselines, VLogger and VideoDirectorGPT, require around 6K and 400 GPU hours, respectively.
Notably, our method still outperforms these baselines even without RAG and TTF (Table \ref{tab: ablations}), with only spatial-temporal regional attention involved.

\section{Evaluation Metrics}
\label{appendix metrics}
We provide detailed evaluation metrics for assessing generated storytelling videos across multiple dimensions in Section 4.1.
\begin{itemize}[noitemsep, topsep=4pt, leftmargin=*]
\item \textbf{Similarity to Reference Images}: We assess the alignment between the generated videos and reference images using Image-Image CLIP~\cite{hessel2021clipscore} and DINO~\cite{oquab2023dinov2} scores. The evaluation is conducted by averaging the CLIP/DINO similarity scores between each frame of the generated video and the reference images, with CLIP-L14~(336px)~\cite{hessel2021clipscore} and DINOv2-B16~\cite{oquab2023dinov2} as image encoders.

\item \textbf{Similarity to Full Narration per Scene}: We measure the alignment between the full narration and the generated videos for each scene using Image-Text CLIP~\cite{hessel2021clipscore} and Video-Text ViCLIP~\cite{wang2023internvid} scores. For the CLIP score, we uniformly sample eight frames from the single-scene video and compute the average score between each frame and the full narration. For the ViCLIP score, we directly use the alignment score between the video and the narration. Per-scene scores are averaged to obtain the overall score for the multi-scene storytelling video.

\item \textbf{Fine-Grained Text Alignment per Scene}: We also assess fine-grained alignment to textual descriptions using Image-Text CLIP~\cite{hessel2021clipscore} and Video-Text ViCLIP~\cite{wang2023internvid} scores. In our story, each narration contains two motions, which we decouple into two consecutive single-motion descriptions using an LLM~\cite{openai_gpt4o}. For each generated single-scene video, we divide it into two segments at the temporal midpoint. We then compute the CLIP/ViCLIP scores between the first segment and the first description and between the second segment and the second description. The average of these two scores provides the per-scene score, and the per-scene scores are further averaged to compute the overall score for the storytelling video.

\item \textbf{Transition Quality}: We assess whether the single-scene video achieves smooth transitions between two motions. To evaluate this, we uniformly sample four frames from the video per scene and calculate the average DINO similarity between adjacent frames. A higher transition score indicates smoother transitions, as it reflects minimal changes in the background across frames.

\item \textbf{Motion Smoothness}: Following~\cite{huang2023vbench}, we utilize motion priors from a video frame interpolation model~\cite{li2023amt} to quantify the smoothness of motion in generated videos (see Supplementary for details), specifically, given one video, we drop the odd-number frames and then use the interpolation model to infer the dropped frames. Then we compute the Mean Absolute Error between the estimated frames and the dropped frames.

\item \textbf{Aesthetic Quality}: We evaluate the artistic and visual appeal of each generated video frame using the LAION aesthetic predictor~\cite{laion_aesthetic_predictor} frame-by-frame. This metric reflects aesthetic aspects such as composition, color richness and harmony, photorealism, naturalness, and overall artistic quality.

\item \textbf{Imaging Quality}: Imaging quality measures distortions such as over-exposure, noise, and blur in generated frames. We evaluate this using the MUSIQ~\cite{ke2021musiq} image quality predictor frame-by-frame, which is trained on the SPAQ dataset~\cite{fang2020cvpr}.
\end{itemize}

\section{Visual Quality Evaluation for SVG}
\label{fullmetrics}

We evaluate the generated videos using a comprehensive set of visual quality and consistency metrics, including aesthetics, imaging quality, motion smoothness, temporal flickering, subject consistency, and background consistency, as shown in Table~\ref{table:visual_quality_extended} and Table~\ref{tab:ablations_quality_only}. Our method achieves the best overall performance across all metrics, demonstrating its effectiveness in producing high-quality and consistent videos. Moreover, the ablation results indicate that introducing the proposed RAG and SR3AI modules does not harm the visual quality, as the full model maintains or improves performance compared to its variants.

\begin{figure*}[p]
    \centering
    \includegraphics[width=1.0\linewidth]{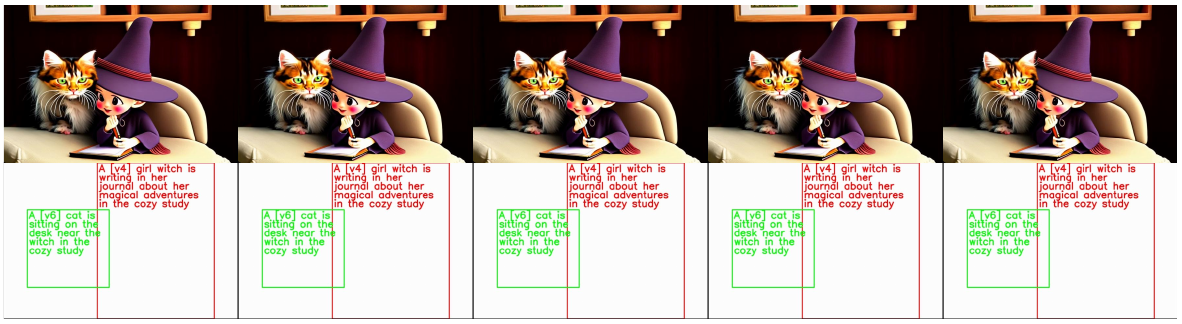}
    \caption{Generated multi-character videos with slightly overlapping regions.}
    \label{fig:overlap}
\end{figure*}

\begin{table*}[p]
\setlength{\tabcolsep}{6pt}
\renewcommand{\arraystretch}{1.2}
\centering
\resizebox{1.0\linewidth}{!}{
\begin{tabular}{lccccccc}
\Xhline{1.0pt}
\textbf{Method} & \textbf{Aesthetics} & \textbf{Imaging} & \textbf{Smoothness} & \textbf{Temp. Flickering} & \textbf{Subject Consistency} & \textbf{BG Consistency} & \textbf{Overall} \\
\hline
VideoDirectorGPT~\cite{lin2023videodirectorgpt} & 42.3 & 60.3 & 94.3 & 91.5 & 75.6 & 87.8 & 75.3 \\
VLogger~\cite{zhuang2024vlogger} & 43.4 & 61.2 & 96.2  & 95.7 & 84.1 & 91.0 & 78.6 \\
\textbf{\ours{} (Ours)} & 
\textbf{55.4} {\small{\color{blue}(+27.65\%)}} & 
\textbf{62.1} {\small{\color{blue}(+1.47\%)}} & 
\textbf{98.1} {\small{\color{blue}(+1.98\%)}} & 
\textbf{96.2} {\small{\color{blue}(+0.52\%)}} & 
\textbf{91.2} {\small{\color{blue}(+8.44\%)}} & 
\textbf{92.3} {\small{\color{blue}(+1.43\%)}} & 
\textbf{82.6} {\small{\color{blue}(+5.03\%)}} \\
\Xhline{1.0pt}
\end{tabular}
}
\vspace{-0.05in}
\caption{\textbf{Evaluation of visual quality and consistency.} Metrics include aesthetics, imaging, motion smoothness, temporal stability, subject/background consistency, and overall quality. \ours{} Achieves the best quality compared with others.}
\label{table:visual_quality_extended}
\end{table*}

\begin{table*}[p]
\setlength{\tabcolsep}{4pt}
\renewcommand{\arraystretch}{1.2}
\centering
\resizebox{1.0\linewidth}{!}{
\begin{tabular}{ccccccccc}
\Xhline{1.0pt}
\textbf{RAG} & \textbf{SR3AI} & \textbf{Aesthetics} & \textbf{Imaging} & \textbf{Smoothness} & \textbf{Temp. Flickering} & \textbf{Subject Consistency} & \textbf{BG Consistency} & \textbf{Overall} \\
\hline
$\times$ & $\times$ & 54.3 & 61.3 & 94.3 & 95.9 & 91.0 & 92.2 & 81.50 \\
$\times$ & \checkmark & 55.4 & 61.9 & 98.0 & 96.3 & 90.9 & 92.3 & 82.47 \\
\checkmark & $\times$ & \textbf{55.6} & 61.9 & 98.1 & 96.0 & 91.1 & 92.0 & 82.45 \\
\checkmark & \checkmark & 55.4 & \textbf{62.1} & \textbf{98.1} & \textbf{96.2} & \textbf{91.2} & \textbf{92.3} & \textbf{82.55} \\
\Xhline{1.0pt}
\end{tabular}
}
\caption{\textbf{Ablation on quality dimensions.} Overall is the mean of all six metrics. Adding RAG and SR3AI doesn't harm the overall visual quality of the backbone model. The last row is our full method \ours{}.}
\label{tab:ablations_quality_only}
\vspace{-10pt}
\end{table*}

\section{Compositional T2V Examples}
\label{t2vcomqual}
In this section, we present qualitative examples demonstrating the capabilities of \ours{} in compositional text-to-video generation. \ours{} effectively generates videos with accurate action binding to different characters, consistent attributes across objects, dynamic attribute changes, motion control, object interactions, and appropriate spatial relationships between objects.  Note that in the overlapped regions, the caption is a merge of all involved captions. For cleaner visualization, we don’t show the merged caption in the layout plans.

\begin{figure*}[t]
  \centering
    \includegraphics[width=0.96\textwidth]{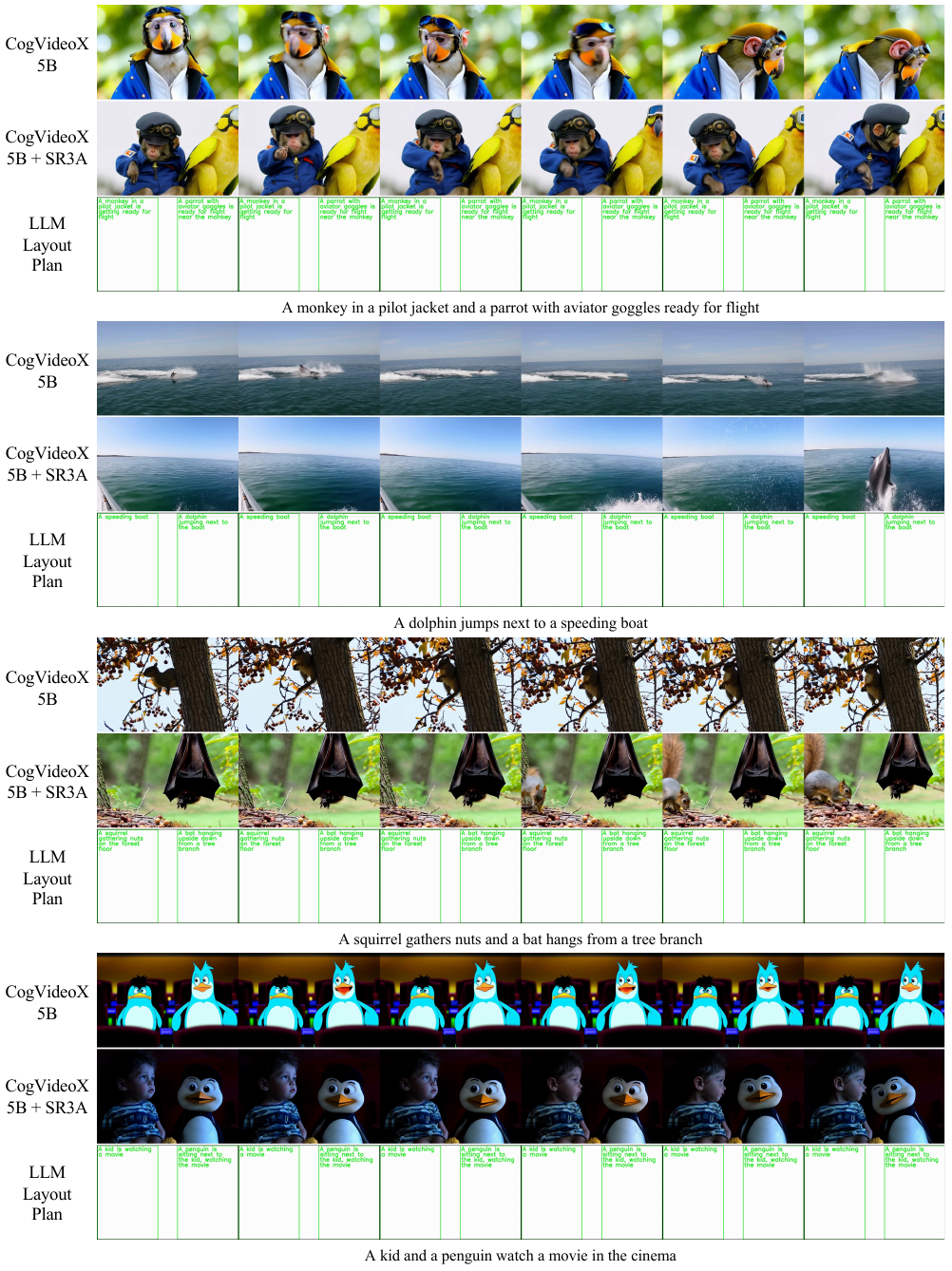}
  \vspace{-0.1in}\caption{
  Qualitative results of \ours{} generated with prompts characterizing {\color{orange}\textbf{action binding}}. SR3A denotes our spatial-temporal region-based attention module.
  }
  \label{fig:action_binding}
  \vspace{-10pt}
\end{figure*}

For action binding, as shown in Figure~\ref{fig:action_binding}, \ours{} generates distinct motions for two objects (e.g., A monkey in a pilot jacket and a parrot with aviator goggles ready for flight), ensuring the actions are correctly bound to their respective objects without interference. 

\begin{figure*}[t]
  \centering
    \includegraphics[width=0.96\textwidth]{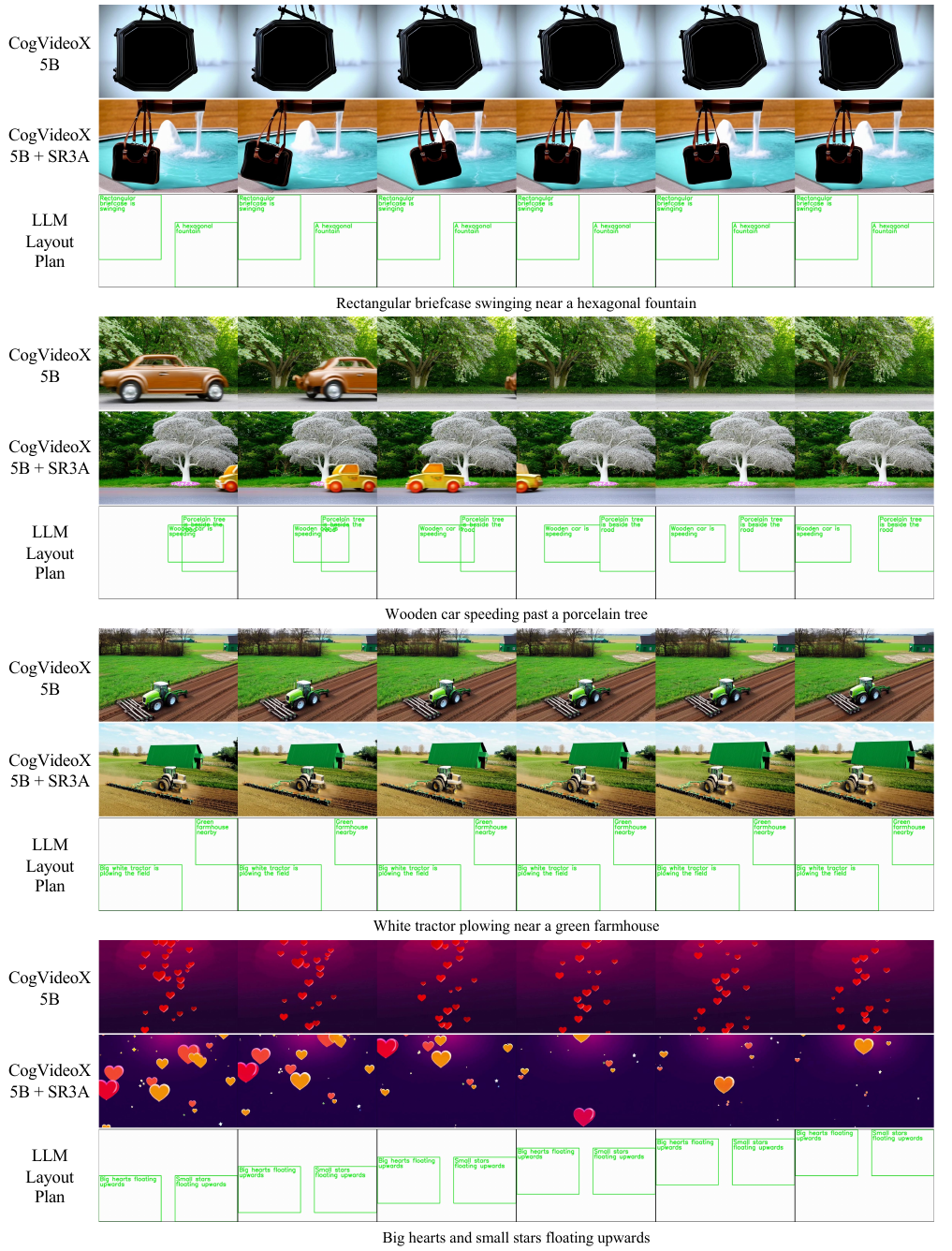}
  \vspace{-0.1in}\caption{
  Qualitative results of \ours{} generated with prompts characterizing {\color{orange}\textbf{consistent attribute binding}}. SR3A denotes our spatial-temporal region-based attention module.
  }
  \label{fig:consistent_attribute_binding}
  \vspace{-10pt}
\end{figure*}

For consistent attribute binding, as illustrated in Figure~\ref{fig:consistent_attribute_binding}, our approach maintains separate attributes for different objects (e.g., Rectangular briefcase swinging near a hexagonal fountain) without any overlap or inconsistency.

\begin{figure*}[t]
  \centering
    \includegraphics[width=0.96\textwidth]{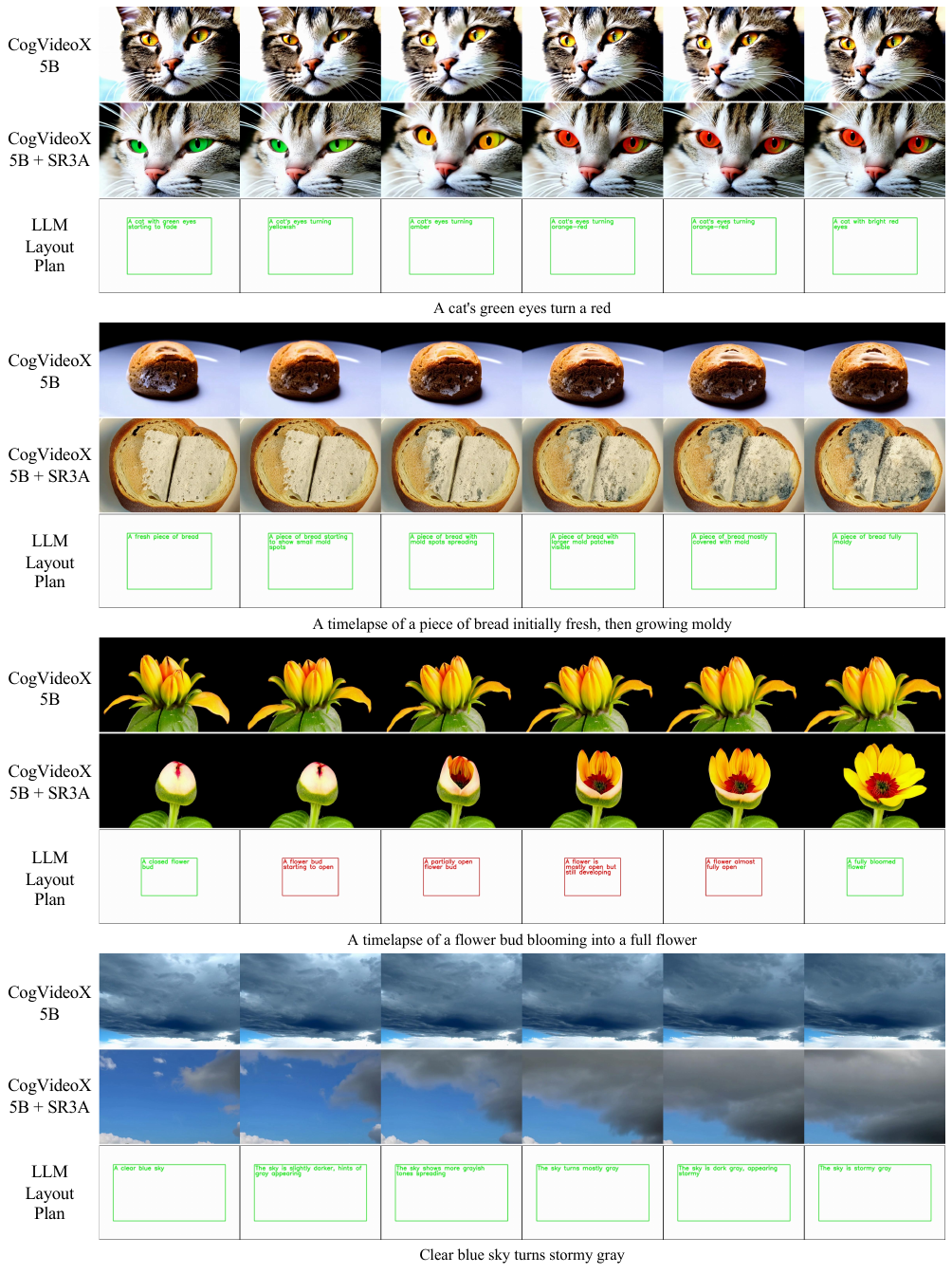}
  \vspace{-0.1in}\caption{
Qualitative results of \ours{} generated with prompts characterizing {\color{orange}\textbf{dynamic attribute binding}}. SR3A denotes our spatial-temporal region-based attention module.
}
  \label{fig:dynamic_attribute_binding}
  \vspace{-10pt}
\end{figure*}

For dynamic attribute changes, as shown in Figure~\ref{fig:dynamic_attribute_binding}, \ours{} naturally transitions object attributes over time (e.g., A timelapse of a piece of bread initially fresh, then growing moldy).

\begin{figure*}[t]
  \centering
    \includegraphics[width=0.96\textwidth]{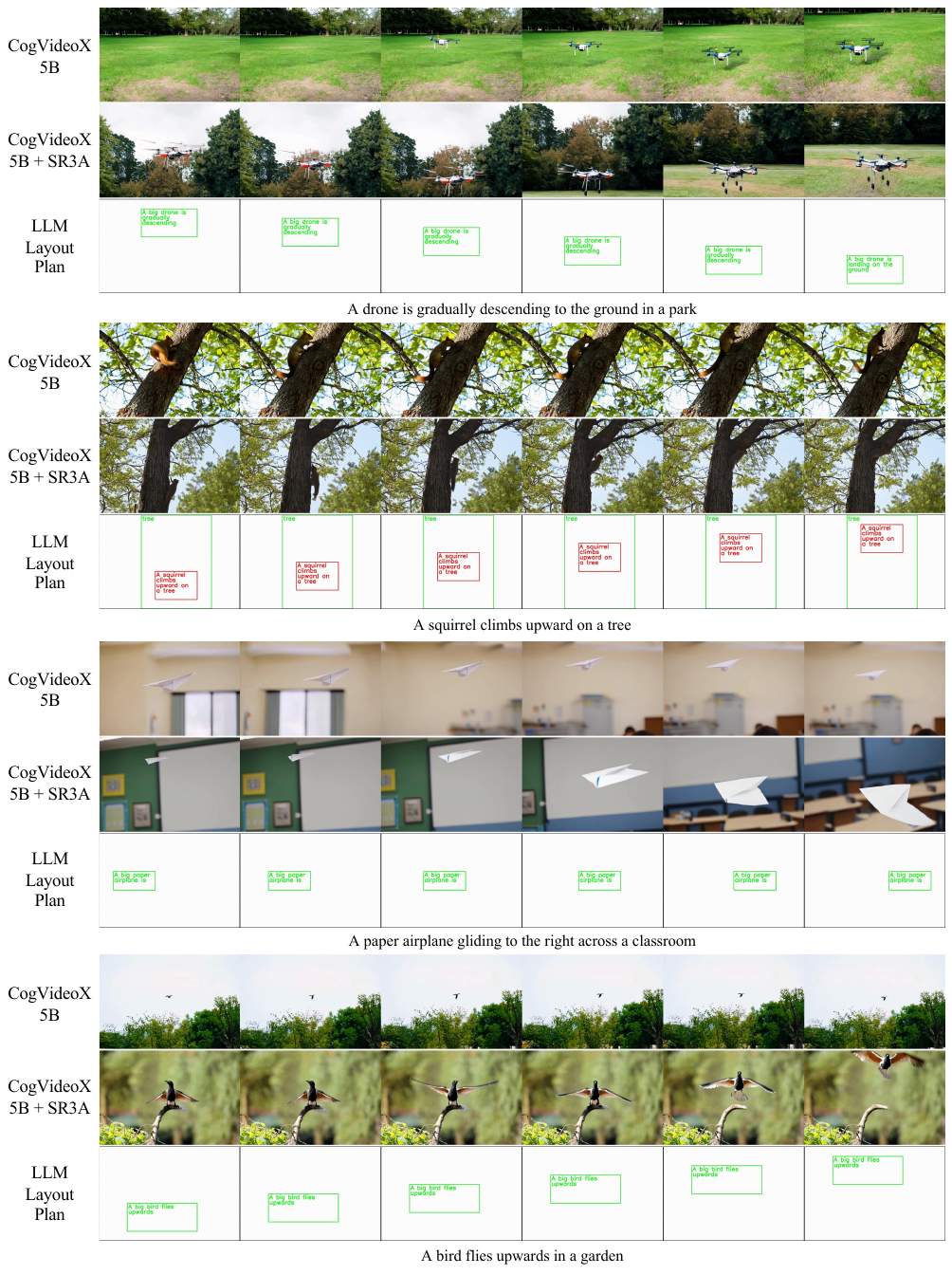}
  \vspace{-0.1in}\caption{
    Qualitative results of \ours{} generated with prompts characterizing {\color{orange}\textbf{motion binding}}. SR3A denotes our spatial-temporal region-based attention module.
  }
  \label{fig:motion_binding}
  \vspace{-10pt}
\end{figure*}

For motion control, as depicted in Figure~\ref{fig:motion_binding}, \ours{} successfully directs object movements in different trajectories (e.g., A squirrel climbs upward on a tree and A drone is gradually descending to the ground in a park).

\begin{figure*}[t]
  \centering
    \includegraphics[width=0.96\textwidth]{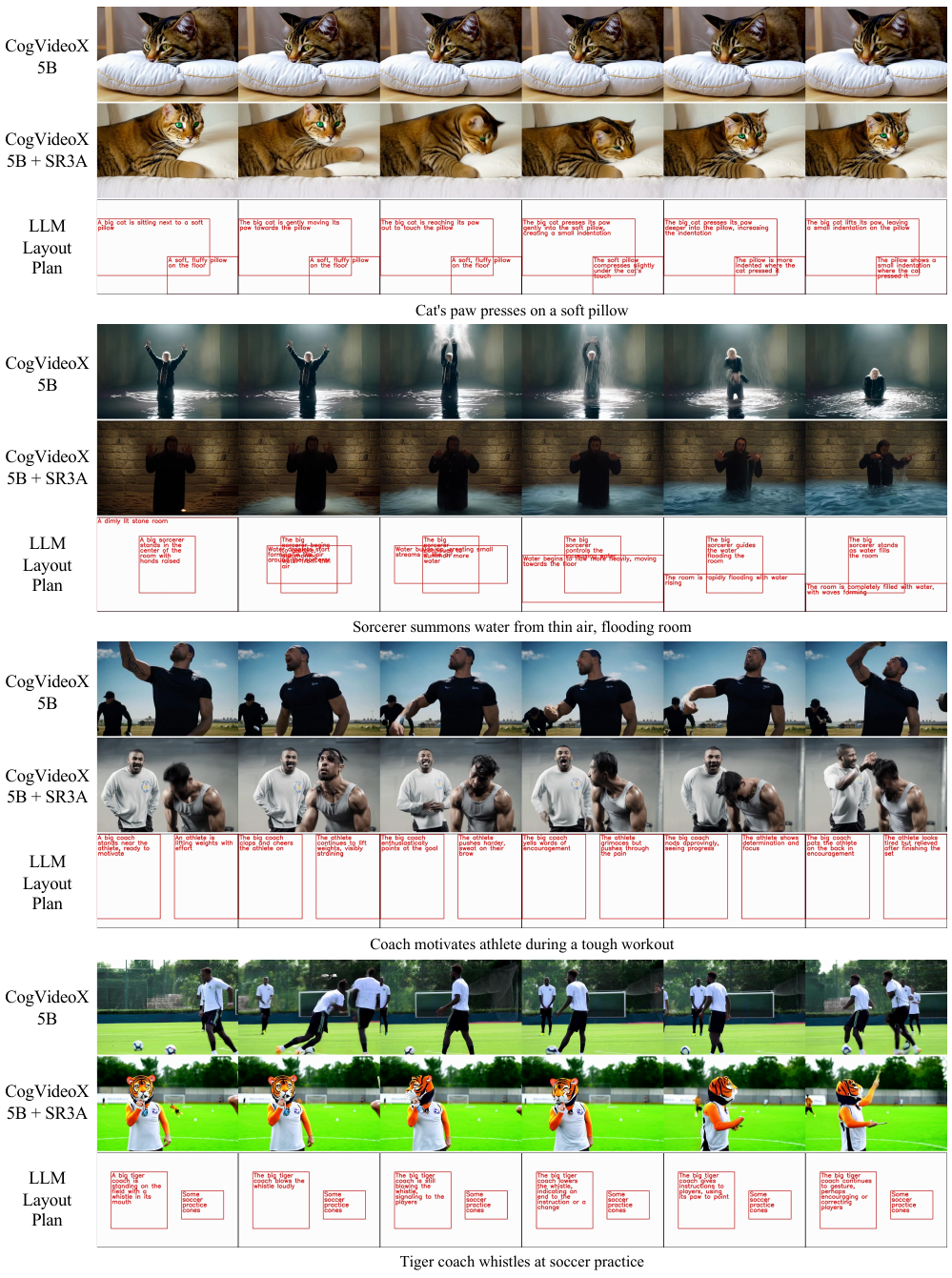}
  \vspace{-0.1in}\caption{
    Qualitative results of \ours{} generated with prompts characterizing {\color{orange}\textbf{object interactions}}. SR3A denotes our spatial-temporal region-based attention module.
  }
  \label{fig:object_interactions}
  \vspace{-10pt}
\end{figure*}

For object interactions, as illustrated in Figure~\ref{fig:object_interactions}, interactions are accurately modeled, adhering to the real world rules (e.g., Cat's paw presses on a soft pillow).

\begin{figure*}[t]
  \centering
    \includegraphics[width=0.96\textwidth]{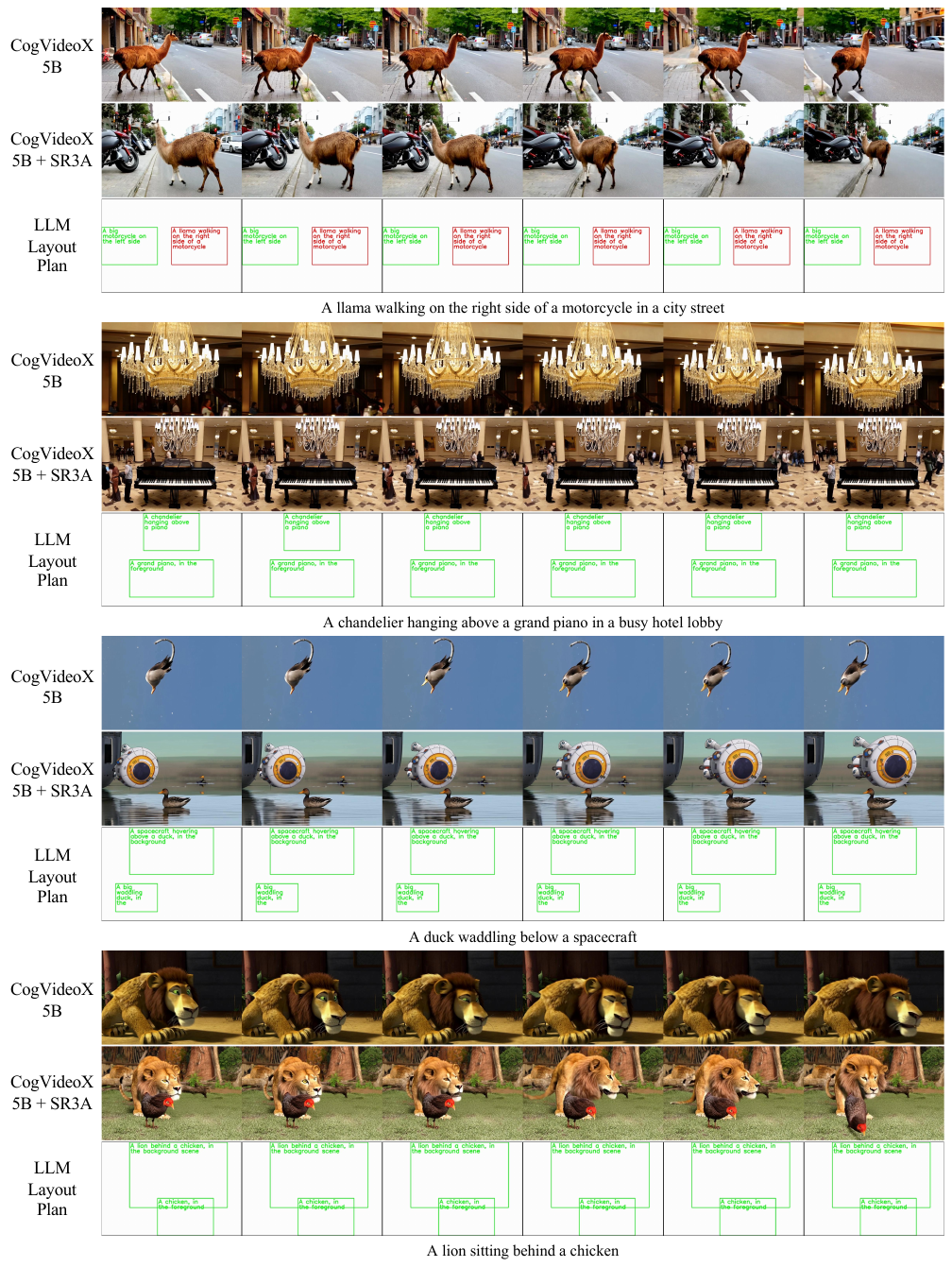}
  \vspace{-0.1in}\caption{
      Qualitative results of \ours{} generated with prompts characterizing {\color{orange}\textbf{spatial relationships}}. SR3A denotes our spatial-temporal region-based attention module.
  }
  \label{fig:spatial_relationships}
  \vspace{-10pt}
\end{figure*}

Finally, for spatial relationships, as shown in Figure~\ref{fig:spatial_relationships}, \ours{} generates rare or imaginative configurations (e.g., a duck positioned under a spacecraft) while maintaining spatial coherence.

These examples highlight the strong performance of \ours{} in generating high-quality compositional text-to-video outputs.

\section{Characters}
\begin{figure*}[t]
  \centering
    \includegraphics[width=1.0\textwidth]{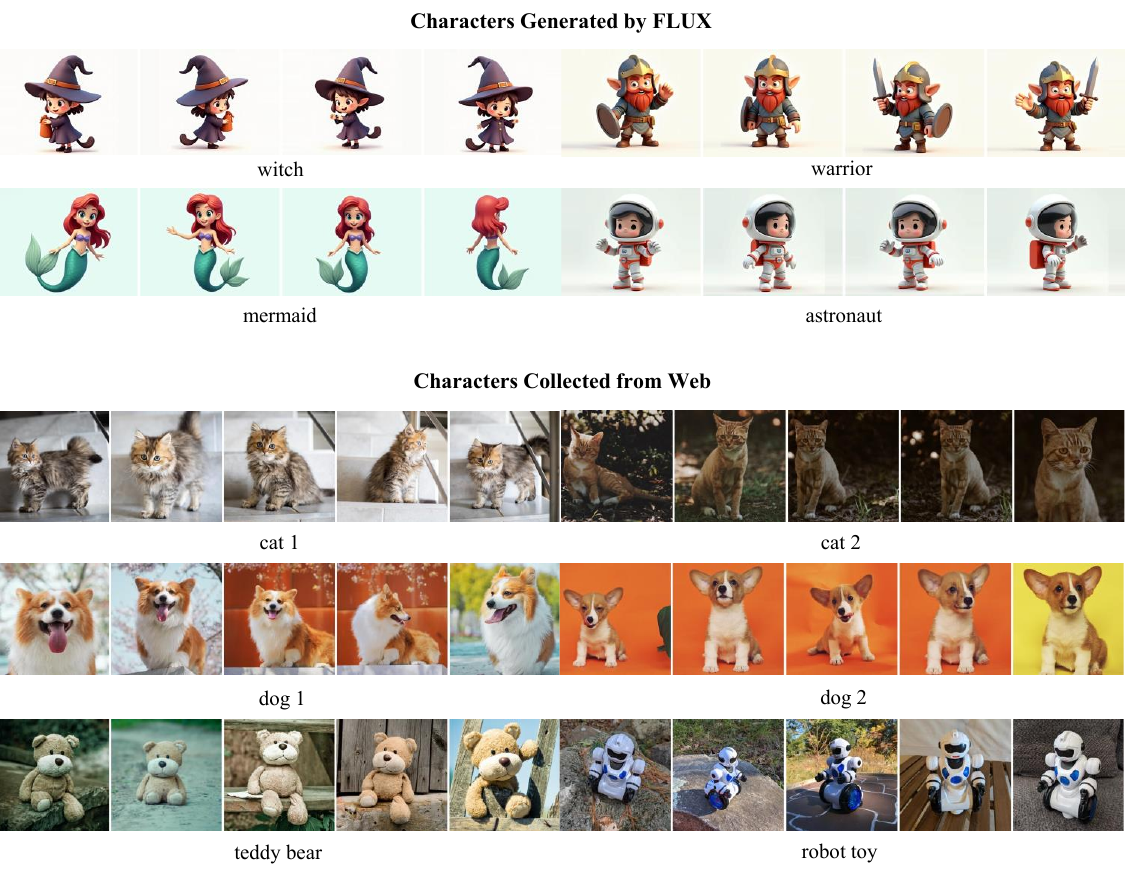}
  \vspace{-0.2in}\caption{
         Qualitative results of \ours{} generated with {\color{orange}\textbf{a single character (mermaid)}}.
  }
  \label{fig:char}
  \vspace{-10pt}
\end{figure*}

We provide character examples in Figure \ref{fig:char}, where the first four are generated using FLUX~\cite{flux}, and the others are collected from existing customization datasets  (CustomConcept101~\citep{kumari2023multi} and Dreambooth~\citep{ruiz2023dreambooth}).

\section{Single-Character Examples}
\begin{figure*}[t]
  \centering
    \includegraphics[width=1.0\textwidth]{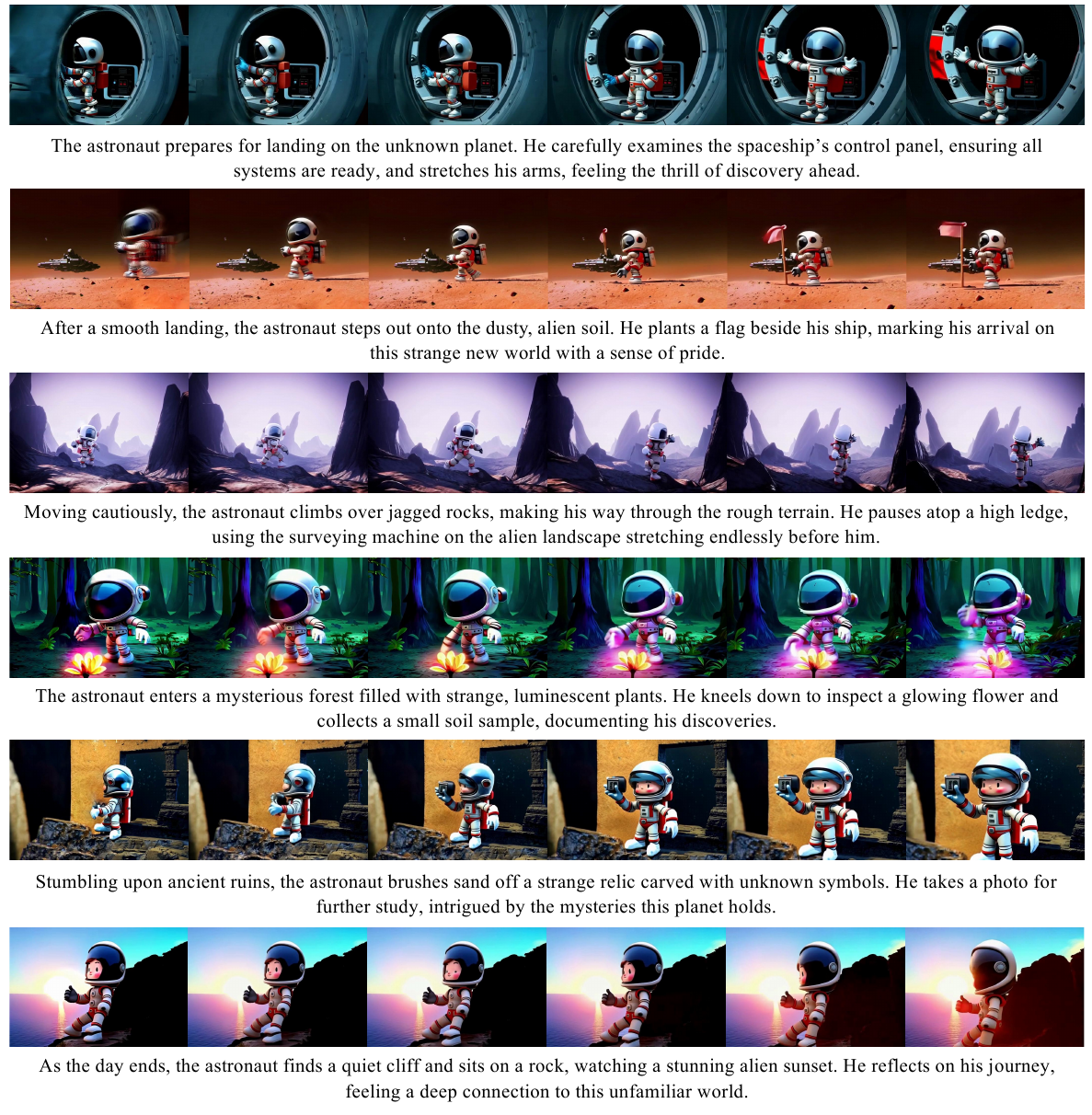}
  \vspace{-0.2in}\caption{
         Qualitative results of \ours{} generated with {\color{orange}\textbf{a single character (astronaut)}}.
  }
  \label{fig:astronaut}
  \vspace{-10pt}
\end{figure*}

\begin{figure*}[t]
  \centering
    \includegraphics[width=1.0\textwidth]{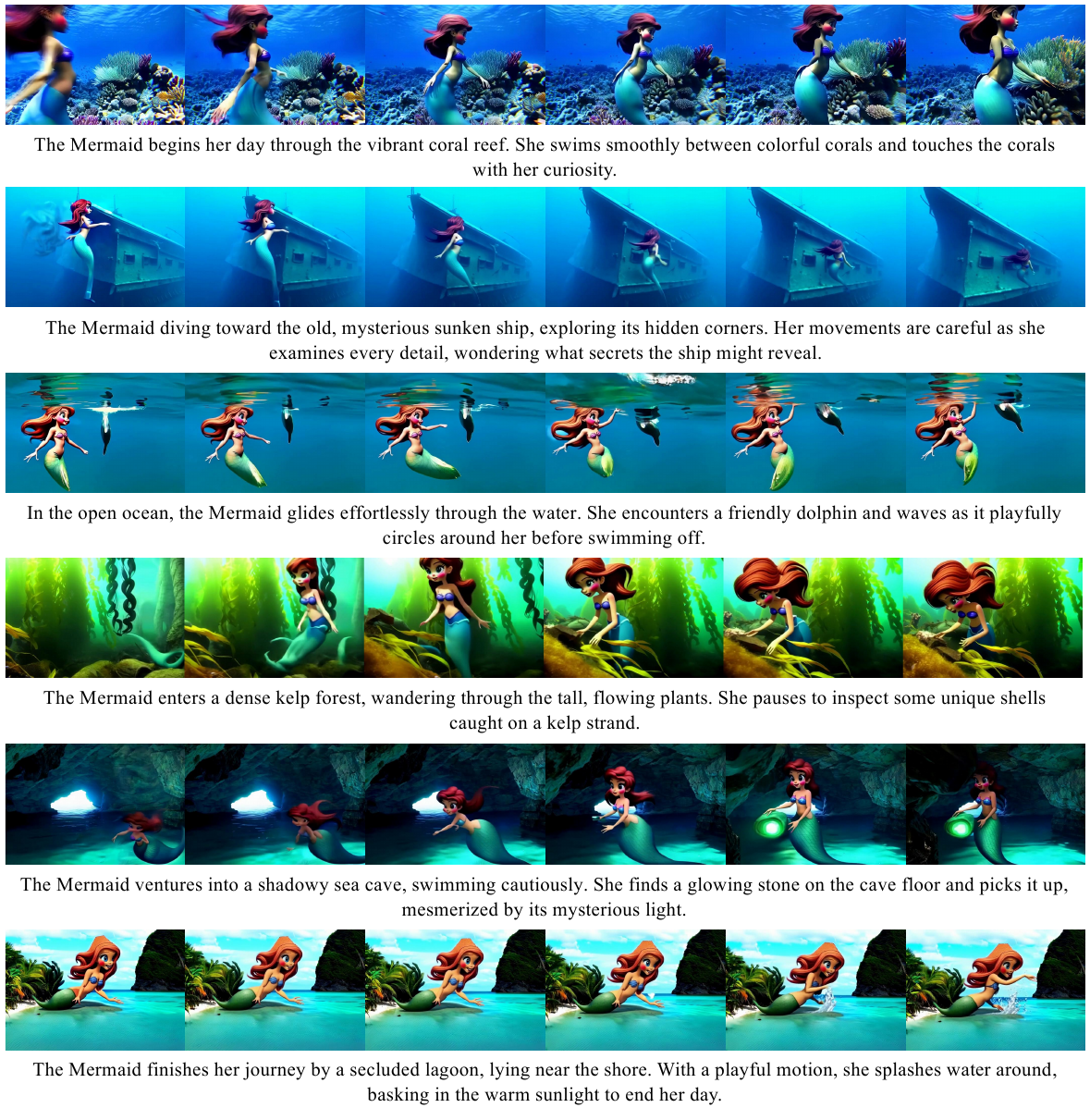}
  \vspace{-0.2in}\caption{
         Qualitative results of \ours{} generated with {\color{orange}\textbf{a single character (mermaid)}}.
  }
  \label{fig:mermaid}
  \vspace{-10pt}
\end{figure*}

In this section, we present qualitative examples of video generation featuring a single main character. As shown in Figure~\ref{fig:mermaid} and Figure~\ref{fig:astronaut}, \ours{} generates consistent characters throughout the entire story. Additionally, in each scene, \ours{} effectively captures multiple events, such as the mermaid first wandering through the plants and then examining unique shells.

\section{Multi-Character Examples}

\begin{figure*}[t]
  \centering
    \includegraphics[width=1.0\textwidth]{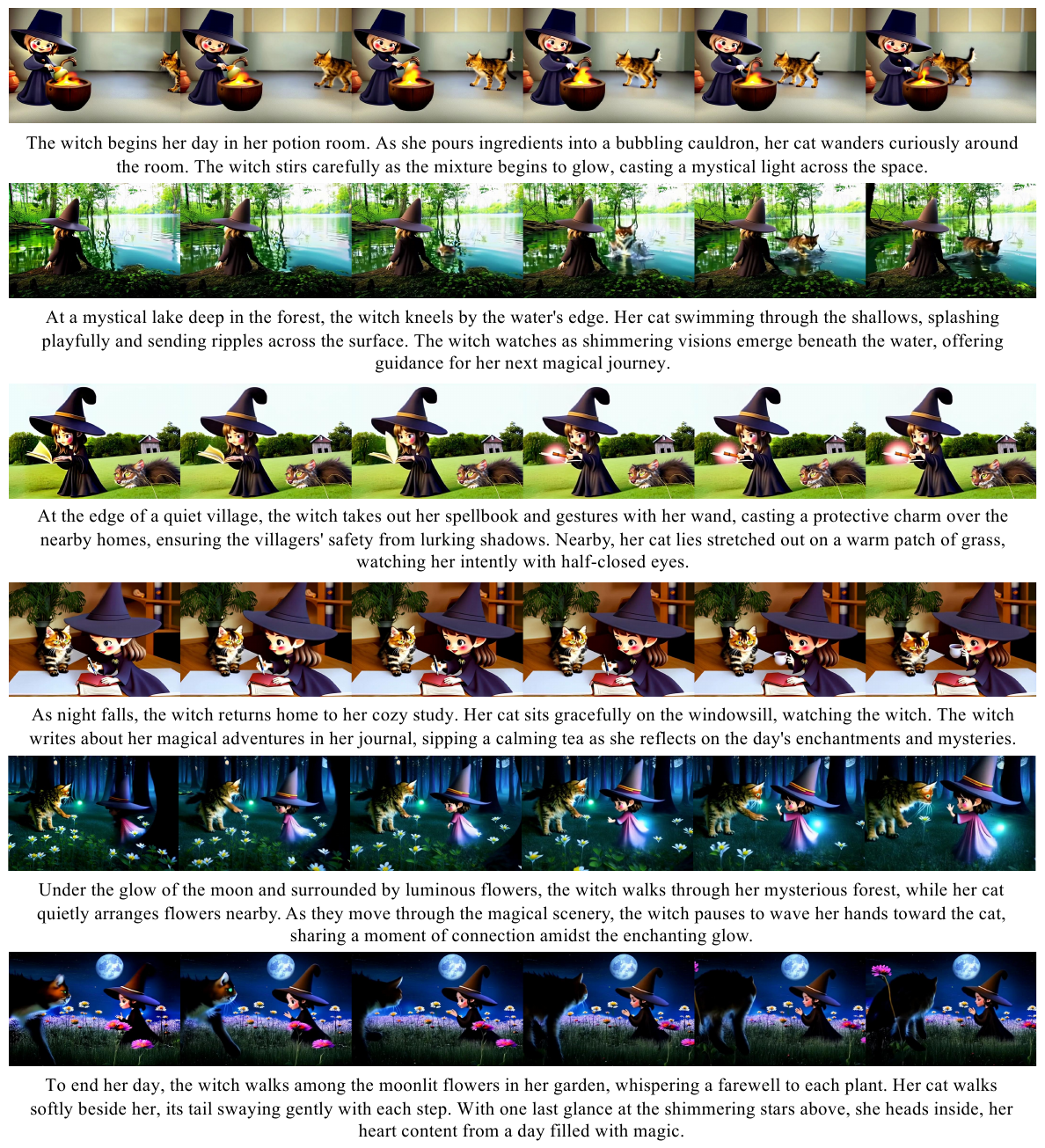}
  \vspace{-0.2in}\caption{
        Qualitative results of \ours{} generated with {\color{orange}\textbf{multiple characters (witch and cat 1)}}.
  }
  \label{fig:witch_and_cat}
  \vspace{-10pt}
\end{figure*}

\begin{figure*}[t]
  \centering
    \includegraphics[width=1.0\textwidth]{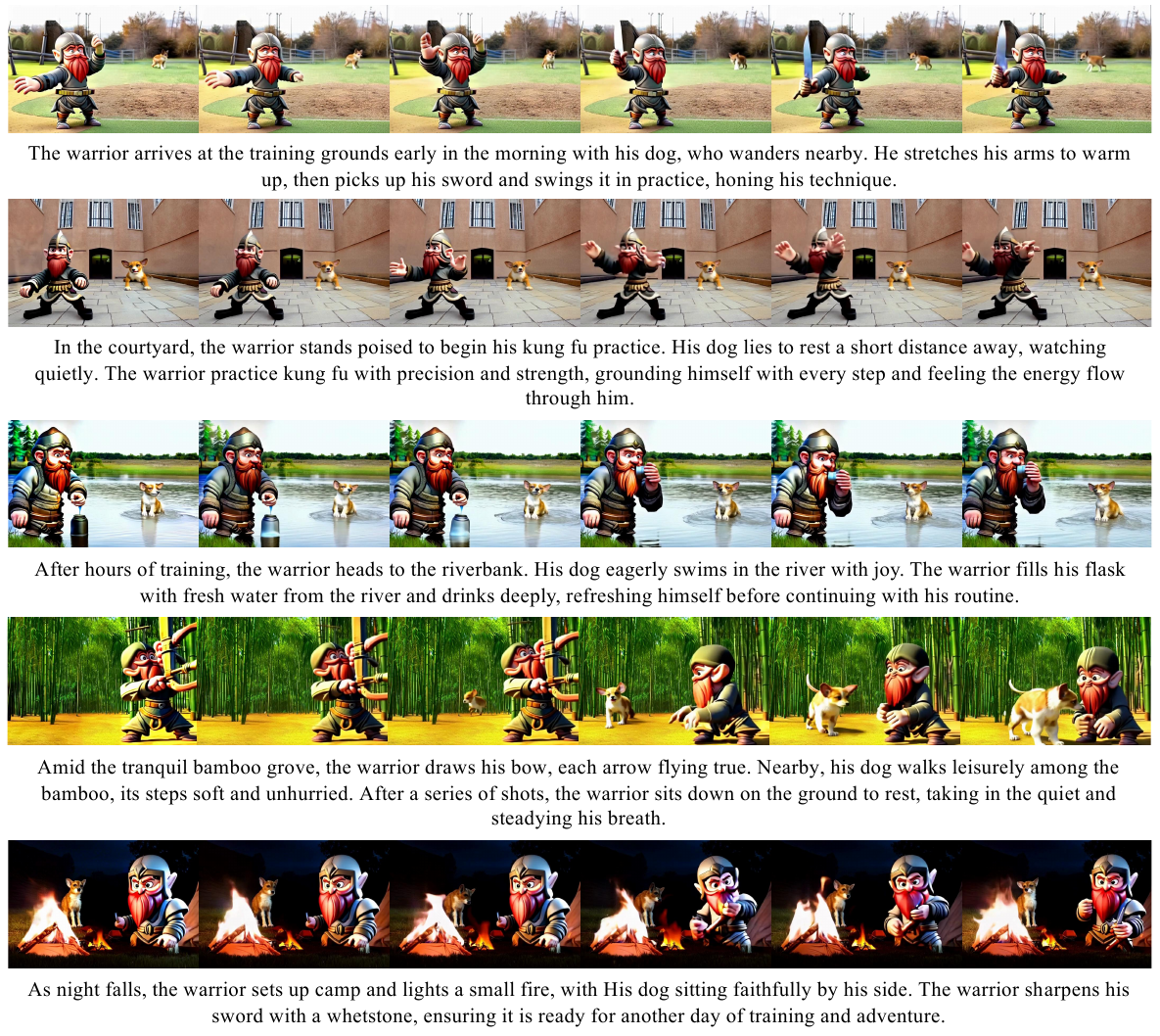}
  \vspace{-0.2in}\caption{
         Qualitative results of \ours{} generated with {\color{orange}\textbf{multiple characters (warrior and dog 2)}}.
  }
  \label{fig:warrior_and_dog}
  \vspace{-10pt}
\end{figure*}

In this section, we present qualitative examples of video generation featuring multiple characters. As illustrated in Figure~\ref{fig:witch_and_cat} and Figure~\ref{fig:warrior_and_dog}, \ours{} generates multi-scene, multi-character videos where each character retains its own motion and interacts seamlessly with others, without any interference. For instance, in Figure~\ref{fig:witch_and_cat}, the witch is shown pouring ingredients while the cat wanders around the room. Even as the cat approaches the witch, their motions remain independent, and the appearance of both characters is consistently preserved throughout the scene.

\section{LLM Prompts}
\label{llm prompts}
We provide detailed LLM prompts for both high-level plans and fine-grained plans in Listing \ref{lst:hlplan} and Listing \ref{lst:fgplan}. respectively. For high-level plans, we use a simple in-context example with instructions, while fine-grained plans require reasoning before generating the output. Example outputs are shown in Listings \ref{lst:hleg} and \ref{lst:fgeg}. We also provide our example input output for merging captions in overlapped regions in Listing ~\ref{lst:overlap}.
For multi-character scenarios, we adapt character-specific words and examples, limit required motions to a maximum of four, and enforce non-overlapping spatial regions for each character.

\section{Multi-Character Video Generation with Overlapping Regions}

We show an example of \ours{} handling overlapped regions in a multi-character setting (see Fig.~\ref{fig:overlap}). In this case, the bounding boxes of the cat and the witch slightly overlap, leading to visual overlap between the characters themselves. This demonstrates \ours{}'s ability to manage such interactions.

\section{Limitation and Future Work}
\textbf{Limitations.} \ours{} build s on a diffusion-based video generation framework and trains lightweight adapters on frozen backbones. As such, its performance and output quality are inherently constrained by the capabilities of the underlying backbone models. If the backbone struggles with rare compositions, complex motions, or fine-grained details, \ours{} may inherit these weaknesses.

\noindent\textbf{Future work.} 
Future improvements may come from adopting stronger and more generalizable backbone models to enhance the overall visual quality. Additionally, advanced LoRA-merging techniques such as CLoRA~\cite{meral2024clora} can be explored to better coordinate multiple character streams and improve quality in overlapping regions. More broadly, developing architectures with stronger compositional reasoning and multi-entity awareness remains an important direction.

\clearpage
\onecolumn

\begin{lstlisting}[caption={The LLM prompt for high-level planning (Section 3.1, Story-Level Croase-Grained Planning). User may input the high-level planning to generate at the \texttt{[Input]} highlighted in \textcolor{Blue}{\textbf{blue}}. We highlight \texttt{\color{Purple}scene}, \texttt{\color{Green}motions}, \texttt{\color{Red}narrations} using same colors as Section 3.1.}, label={lst:hlplan}]
Consider you are an expert in writing stories. I will provide you with a topic, and you need to create a multi-scene story with 5 to 8 scenes. Each scene should describe the events taking place, emphasizing highlighting human actions or motions. Limit each scene to a maximum of 2 distinct human motions. Your output should include both the scene and motions as well as the narration, where the motion should be in present progressive.

[Input] Teddy's one day

[Output]
(*@\textcolor{Purple}{Scene 1}@*): bedroom
(*@\textcolor{Green}{Motions}@*):
waking up, stretching arms
(*@\textcolor{Red}{Narration}@*):
Teddy starts his day in his bedroom. He wakes up from the bed and stretches his arms beside it, getting ready for a new day.

(*@\textcolor{Purple}{Scene 2}@*): kitchen
(*@\textcolor{Green}{Motions}@*):
pouring cereal, drinking
(*@\textcolor{Red}{Narration}@*):
Teddy is in the kitchen for breakfast. He pours cereal into a bowl in the kitchen, and drinks the soup, preparing his morning meal.

(*@\textcolor{Purple}{Scene 3}@*): backyard
(*@\textcolor{Green}{Motions}@*):
watering plants, examining flower
(*@\textcolor{Red}{Narration}@*):
Teddy spends some time outside in the backyard. He waters the garden plants and kneels down to examine a flower, enjoying the fresh air and tending to the plants.

(*@\textcolor{Purple}{Scene 4}@*): park
(*@\textcolor{Green}{Motions}@*):
jogging, breathing
(*@\textcolor{Red}{Narration}@*):
Teddy visits the park for some exercise. He jogs in the park, then he becomes very tired so he breaths heavily, regaining his energy.

(*@\textcolor{Purple}{Scene 5}@*): forest
(*@\textcolor{Green}{Motions}@*):
hiking, sitting
(*@\textcolor{Red}{Narration}@*):
Teddy goes hiking on a forest trail. He hikes up the trail and then sits on a rock to rest, appreciating the tranquility of nature.

(*@\textcolor{Purple}{Scene 6}@*): living room
(*@\textcolor{Green}{Motions}@*):
sitting, reclining
(*@\textcolor{Red}{Narration}@*):
Teddy winds down at the end of the day in the living room. He sits on the couch and then reclines to relax, ending his day in comfort and peace.

[Input] (*@\textcolor{Blue}{\textbf{Mermaid's Adventure}}@*)
\end{lstlisting}

\clearpage

\begin{lstlisting}[caption={The example of the output from the high-level planning prompt, from Listing \ref{lst:hlplan}. We highlight \texttt{\color{Purple}scene}, \texttt{\color{Green}motions}, \texttt{\color{Red}narrations} using same colors as Section 3.1.}, label={lst:hleg}]
[Output]
(*@\textcolor{Purple}{Scene 1}@*): Coral Reef
(*@\textcolor{Green}{Motions}@*):
swimming, touching
(*@\textcolor{Red}{Narration}@*):
The Mermaid begins her day through the vibrant coral reef. She swims smoothly between colorful corals and touches the corals with her curiosity.

(*@\textcolor{Purple}{Scene 2}@*): Sunken Ship
(*@\textcolor{Green}{Motions}@*):
diving, exploring
(*@\textcolor{Red}{Narration}@*):
The Mermaid diving toward the old, mysterious sunken ship, exploring its hidden corners. Her movements are careful as she examines every detail, wondering what secrets the ship might reveal.

(*@\textcolor{Purple}{Scene 3}@*): Open Ocean
(*@\textcolor{Green}{Motions}@*):
gliding, waving
(*@\textcolor{Red}{Narration}@*):
In the open ocean, the Mermaid glides effortlessly through the water. She encounters a friendly dolphin and waves as it playfully circles around her before swimming off.

(*@\textcolor{Purple}{Scene 4}@*): Kelp Forest
(*@\textcolor{Green}{Motions}@*):
wandering, inspecting
(*@\textcolor{Red}{Narration}@*):
The Mermaid enters a dense kelp forest, wandering through the tall, flowing plants. She pauses to inspect some unique shells caught on a kelp strand.

(*@\textcolor{Purple}{Scene 5}@*): Sea Cave
(*@\textcolor{Green}{Motions}@*):
swimming, picking up
(*@\textcolor{Red}{Narration}@*):
The Mermaid ventures into a shadowy sea cave, swimming cautiously. She finds a glowing stone on the cave floor and picks it up, mesmerized by its mysterious light.

(*@\textcolor{Purple}{Scene 6}@*): Lagoon
(*@\textcolor{Green}{Motions}@*):
lying, splashing
(*@\textcolor{Blue}{Narration}@*):
The Mermaid finishes her journey by a secluded lagoon, lying near the shore. With a playful motion, she splashes water around, basking in the warm sunlight to end her day.
\end{lstlisting}

\newpage
\begin{lstlisting}[caption={LLM prompt for fine-grained frame-level plan (Section 3.1, Scene-Level Fine-Grained Planning) per narration. We highlight \texttt{\textcolor{deepblue}{Background}}, \texttt{\textcolor{deepgreen}{[entity, motion, caption]}} and layout \texttt{\textcolor{brown}{[left, top, right, bottom]}} using same colors as Section 3.1.}, label={lst:fgplan}]
Assuming the frame size is normalized to the range 0-1, you need to give a possible 6-frame layout plan at 1fps with the relevant regions, containing entity with corresponding motion, caption and bounding box involved in the input example motion and narration. You should follow these instructions:

1. [Background and Regions] You need to give a background of the videos, then list all regions with related entity, motion, caption, and bounding box, for each frame.
2. [Bounding Box Size] Each bounding box of the region is one rectangle or square box in the layout, and the size of boxes should be AS LARGE AS POSSIBLE. The width and height of each bounding box should be at least 0.2.
3. [Bounding Box Format] Every region should contain one related motion and caption, with the bounding boxes in the format of [[entity1, motion1, caption1], [left, top, right, bottom]]. If the entity doesn't involve any motion, use "none" as the motion (e.g. ["table", "none", "a table in the forest"]).
4. [Captions and Motions)] "IMPORTANT" For each entity, you should give a caption containing the entity and motion, with the provided background.
5. [Allowing Overlaps for Interaction Contexts] Regions can overlap if necessary, particularly when entities are interacting. For example, in "Teddy is pouring water into a bowl," you can have a region for "a bowl" overlapping or near Teddy's region. Similarly, in "Teddy is sitting by the river," you can have a region for "river bank" overlapping or positioned beneath Teddy's region.
5. [Allowing Overlaps for Interaction Contexts] Regions can overlap if necessary, particularly when entities are interacting. For example, in "Teddy is pouring water into a bowl," you can have a region for "a bowl" overlapping or near Teddy's region. Similarly, in "Teddy is sitting by the river," you can have a region for "river bank" overlapping or positioned beneath Teddy's region.
6. [Interaction with Objects] For an object listed in previous frames, when the entity is interacting with it in the current frames, you can still list the interacted object, then also use a whole region describing the character interacting with the object (e.g. [["rock", "none", "a rock in the forest"], [0.0, 0.8, 0.2, 1.0]] before entity Teddy is interacting with it, [["rock", "none", "a rock in the forest"], [0.0, 0.8, 0.2, 1.0]], [["Teddy", "sitting", "Teddy is sitting on the rock in the forest"], [0.0, 0.0, 0.4, 1.0]] when interacting).
7. [Static Interaction Objects] In most cases, the interacted objects should be static. For example, "Teddy is greeting to people" then "people" should have no bounding box changes. For some non-static objects like "the bottle" of "Teddy is drinking water", ignore such objects to be listed seperatedly.
8. [Motion Limitation] You should not use other motions outside the provided narration. The motion and caption should all be in PRESENT PROGRESSIVE.
9. [Motion Duration] One motion should last at least two frames.
10. [Reasoning] Add reasoning before you generate the region plan, explaining how you will allocate different events/motions to different frames and how the entity will be moving (e.g., left to right or staying static).
11. [Smart Motion Allocation] Allocate motions smartly by reasoning the frames they need (e.g., static motions like standing will require fewer frames but walking may need more). The background should not contain information about the characters (e.g., Teddy's xxx is not allowed).
12. [Common Sense in Layout] Make sure the locations of the generated bounding boxes are consistent with common sense. You need to generate layouts from the close-up camera view of the event. The layout difference between two adjacent frames should not be too large, considering the small interval.
13. [Motions for Large Position Changes] If you want to move the entity largely by changing its bounding boxes (e.g., from most right to most left), make sure the motion naturally involves position changes (e.g., walking, running, flying, riding a bike). Otherwise, avoid big changes in related bounding boxes (e.g., cooking, jumping, playing guitar, etc., where bounding boxes should not change significantly).
14. [Big Regions Preference] We prioritize using large regions for main entities. For example, [0.2, 0.0, 0.8, 1.0] for a main teddy bear, as long as it fits the scene. Do not use small regions (like with only 0.2~0.3 width and height) as possible as you can.
15. [Motion Caption Consistency] Ensure that each caption accurately reflects the stated motion. For example, if the motion is "walking" the caption should match this exactly, such as "Teddy is walking in the park." Avoid any inconsistencies where the motion described in the caption does not match the stated motion, such as "Teddy is running in the park" when the motion is "walking." This consistency maintains clarity and alignment with the narration.


Use format: 
*Reasoning*
reason

*Plan*
(*@\textcolor{deepblue}{Background: background}@*)
Frame_1: [(*@\textcolor{deepgreen}{[entity1, motion1, caption1]}@*), (*@\textcolor{brown}{[left, top, right, bottom]}@*)], [(*@\textcolor{deepgreen}{[entity2, motion2, caption2]}@*), (*@\textcolor{brown}{[left, top, right, bottom]}@*)], ..., [(*@\textcolor{deepgreen}{[entity3, motion3, caption3]}@*), (*@\textcolor{brown}{[left, top, right, bottom]}@*)]
Frame_2: [(*@\textcolor{deepgreen}{[entity1, motion1, caption1]}@*), (*@\textcolor{brown}{[left, top, right, bottom]}@*)], [(*@\textcolor{deepgreen}{[entity2, motion2, caption2]}@*), (*@\textcolor{brown}{[left, top, right, bottom]}@*)], ..., [(*@\textcolor{deepgreen}{[entity3, motion3, caption3]}@*), (*@\textcolor{brown}{[left, top, right, bottom]}@*)]
...
Frame_6: [(*@\textcolor{deepgreen}{[entity1, motion1, caption1]}@*), (*@\textcolor{brown}{[left, top, right, bottom]}@*)], [(*@\textcolor{deepgreen}{[entity2, motion2, caption2]}@*), (*@\textcolor{brown}{[left, top, right, bottom]}@*)], ..., [(*@\textcolor{deepgreen}{[entity3, motion3, caption3]}@*), (*@\textcolor{brown}{[left, top, right, bottom]}@*)]

Reasoning: ...

Example 1:
[Input]
Motion: 
walking, sitting
Narration:
Teddy goes to a forest. He walks on the trail and then sits on a rock to rest, appreciating the tranquility of nature.

[Output]
*Reasoning*
Listed motions are: walking, sitting.
Related entities and motions: Teddy (from walking to sitting), rock (will be sit by Teddy in the end, but no related motions).
Motion frames allocation: The main entity involve motion changes is Teddy. Teddy's motion changes from walking to sitting. As sitting takes less time, we should allocate more frames to walking. so the plan is 4 frames for walking and 2 frames for sitting. (Following bullet point 11 [Smart Motion Allocation])
Bounding box changes: For Teddy, the first motion is walking, which involves position changes a little bit. And the second motion is sitting, which should be a bounding-box-static motion. So the Teddy bear can have bounding boxes changes at first and finally interacts with the  For the rock, it doesn't involve any motion, so it should be static from far from to close to, and finally interacting with Teddy. (Following bullet point 13 [Motions for Large Position Changes])
Interaction: This input narration involve interaction between the rock and Teddy and happens in late frames, so the rock and Teddy will have one merged-region where the entity is Teddy and the caption is about Teddy sitting on the rock. ALso the rock will also be listed from frame 1 to 3, while maintained from 4 to 6, as a static object. (Following bullet point 5 [Allowing Overlaps for Interaction Contexts], 6 [Interaction with Objects], 7 [Static Interaction Objects])

In conclusion, the plan can be: Teddy is at the right side at the beginning of the video, and there's also a rock on the left right corner. From Frame 1 to Frame 4, the Teddy bear is hiking on the trail moving from right to left to approaching the rock. In Frame 5 and Frame 6, it reaches the rock and sits on the rock (which is on the left corner) to rest, in the left side of the video.

*Plan*
(*@\textcolor{deepblue}{Background: the forest}@*)
Frame_1: [(*@\textcolor{deepgreen}{["Teddy", "hiking", "Teddy is hiking on a trail in the forest"]}@*), (*@\textcolor{brown}{[0.6, 0.0, 1.0, 1.0]}@*)], [(*@\textcolor{deepgreen}{["rock", "none", "a rock in the forest"]}@*), (*@\textcolor{brown}{[0.0, 0.8, 0.2, 1.0]}@*)]
Frame_2: [(*@\textcolor{deepgreen}{["Teddy", "hiking", "Teddy is hiking on a trail in the forest"]}@*), (*@\textcolor{brown}{[0.47, 0.0, 0.87, 1.0]}@*)], [(*@\textcolor{deepgreen}{["rock", "none", "a rock in the forest"]}@*), (*@\textcolor{brown}{[0.0, 0.8, 0.2, 1.0]}@*)]
Frame_3: [(*@\textcolor{deepgreen}{["Teddy", "hiking", "Teddy is hiking on a trail in the forest"]}@*), (*@\textcolor{brown}{[0.33, 0.0, 0.73, 1.0]}@*)], [(*@\textcolor{deepgreen}{["rock", "none", "a rock in the forest"]}@*), (*@\textcolor{brown}{[0.0, 0.8, 0.2, 1.0]}@*)]
Frame_4: [(*@\textcolor{deepgreen}{["Teddy", "hiking", "Teddy is hiking on a trail in the forest"]}@*), (*@\textcolor{brown}{[0.2, 0.0, 0.6, 1.0]}@*)], [(*@\textcolor{deepgreen}{["rock", "none", "a rock in the forest"]}@*), (*@\textcolor{brown}{[0.0, 0.8, 0.2, 1.0]}@*)]
Frame_5: [(*@\textcolor{deepgreen}{["Teddy", "sitting", "Teddy is sitting on the rock in the forest"]}@*), (*@\textcolor{brown}{[0.0, 0.0, 0.4, 1.0]}@*)], [(*@\textcolor{deepgreen}{["rock", "none", "a rock in the forest"]}@*), (*@\textcolor{brown}{[0.0, 0.8, 0.2, 1.0]}@*)]
Frame_6: [(*@\textcolor{deepgreen}{["Teddy", "sitting", "Teddy is sitting on the rock in the forest"]}@*), (*@\textcolor{brown}{[0.0, 0.0, 0.4, 1.0]}@*)], [(*@\textcolor{deepgreen}{["rock", "none", "a rock in the forest"]}@*), (*@\textcolor{brown}{[0.0, 0.8, 0.2, 1.0]}@*)]

[Input]
Motions:
swimming, touching
Narration:
The Mermaid begins her day through the vibrant coral reef. She swims smoothly between colorful corals and touches the corals with her curiosity.
\end{lstlisting}

\newpage
\begin{lstlisting}[caption={The example output of fine-grained planning prompt~\ref{lst:fgplan}. We highlight \texttt{\textcolor{deepblue}{Background}}, \texttt{\textcolor{deepgreen}{[entity, motion, caption]}} and layout \texttt{\textcolor{brown}{[left, top, right, bottom]}} using same colors as Section 3.1.}, label={lst:fgeg}]
[Output]
*Reasoning*
Listed motions are: swimming, touching.
Related entities and motions: Mermaid (from swimming to touching), corals (being touched by the mermaid).
Motion frames allocation: The main entity involving motion changes is the Mermaid. The first motion, swimming, can last relatively longer with more frame coverage as it involves position changes. Touching corals happens afterward and should occupy fewer frames with minimal position changes.
Bounding box changes: The swimming motion involves position changes, so the bounding box for the Mermaid can have slight shifts, swimming through the corals from left to right. The touching motion of the Mermaid should be static, focusing on the interaction at the coral, which is near the center of the frame.
Interaction: The interaction between the Mermaid and the corals happens later in the narration, so a combined region featuring the Mermaid touching the corals will be involved in the last two frames. 
Non-interacting corals will appear in all frames, having stable bounding boxes across the video to show the environment and the Mermaid's interaction context.

The plan is as follows: The Mermaid swims through the vibrant coral reef, starting from near the left, moving towards the center, where the corals are located. On reaching the center, she begins touching the corals with fascination.

*Plan*
(*@\textcolor{deepblue}{Background: the vibrant coral reef full of colors and life}@*)
Frame_1: [(*@\textcolor{deepgreen}{["Mermaid", "swimming", "The Mermaid is swimming through the vibrant coral reef"]}@*), (*@\textcolor{brown}{[0.0, 0.0, 0.4, 1.0]}@*)], [(*@\textcolor{deepgreen}{["corals", "none", "Colorful corals in the reef"]}@*), (*@\textcolor{brown}{[0.5, 0.3, 0.8, 0.6]}@*)]
Frame_2: [(*@\textcolor{deepgreen}{["Mermaid", "swimming", "The Mermaid is swimming through the vibrant coral reef"]}@*), (*@\textcolor{brown}{[0.15, 0.0, 0.55, 1.0]}@*)], [(*@\textcolor{deepgreen}{["corals", "none", "Colorful corals in the reef"]}@*), (*@\textcolor{brown}{[0.5, 0.3, 0.8, 0.6]}@*)]
Frame_3: [(*@\textcolor{deepgreen}{["Mermaid", "swimming", "The Mermaid is swimming through the vibrant coral reef"]}@*), (*@\textcolor{brown}{[0.3, 0.0, 0.7, 1.0]}@*)], [(*@\textcolor{deepgreen}{["corals", "none", "Colorful corals in the reef"]}@*), (*@\textcolor{brown}{[0.5, 0.3, 0.8, 0.6]}@*)]
Frame_4: [(*@\textcolor{deepgreen}{["Mermaid", "swimming", "The Mermaid is swimming through the vibrant coral reef"]}@*), (*@\textcolor{brown}{[0.45, 0.0, 0.85, 1.0]}@*)], [(*@\textcolor{deepgreen}{["corals", "none", "Colorful corals in the reef"]}@*), (*@\textcolor{brown}{[0.5, 0.3, 0.8, 0.6]}@*)]
Frame_5: [(*@\textcolor{deepgreen}{["Mermaid", "touching", "The Mermaid is touching the corals with her curiosity"]}@*), (*@\textcolor{brown}{[0.4, 0.1, 0.7, 0.9]}@*)], [(*@\textcolor{deepgreen}{["corals", "none", "Colorful corals in the reef"]}@*), (*@\textcolor{brown}{[0.5, 0.3, 0.8, 0.6]}@*)]
Frame_6: [(*@\textcolor{deepgreen}{["Mermaid", "touching", "The Mermaid is touching the corals with her curiosity"]}@*), (*@\textcolor{brown}{[0.4, 0.1, 0.7, 0.9]}@*)], [(*@\textcolor{deepgreen}{["corals", "none", "Colorful corals in the reef"]}@*), (*@\textcolor{brown}{[0.5, 0.3, 0.8, 0.6]}@*)]
\end{lstlisting}

\begin{lstlisting}[caption={The example input output for getting merged prompt for overlapping regions. The model input is in black while the output is in blue}, label={lst:overlap}]

Given multiple captions describing the same scene, merge them into a single, coherent caption that includes all key actions or entities, avoids redundancy, and maintains natural fluency.

[Captions]

1. A man is throwing a frisbee in the park.

2. A dog is running toward the frisbee with its tongue out.

[Merged Caption]
(*@\textcolor{blue}{A man throws a frisbee in the park while a dog runs toward it with its tongue out.}@*)
\end{lstlisting}

\twocolumn

\end{document}